\documentclass[10pt,twocolumn,letterpaper]{article}

\usepackage{wacv}
\usepackage{times}
\usepackage{epsfig}
\usepackage[caption=false]{subfig}
\usepackage{graphicx}
\usepackage{amsmath}
\usepackage{amssymb}
\usepackage{multirow}
\usepackage{comment}

\usepackage{xcolor}
\newcommand{\hs}[1]{{\color{red}Hossein: #1}}

\wacvfinalcopy 

\ifwacvfinal
\def\assignedStartPage{1} 
\fi

\ifwacvfinal
\usepackage[breaklinks=true,bookmarks=false]{hyperref}
\else
\usepackage[pagebackref=true,breaklinks=true,colorlinks,bookmarks=false]{hyperref}
\fi

\ifwacvfinal
\else
\fi

\begin{document}

\title{Identification of Attack-Specific Signatures in Adversarial Examples}

\author{

Hossein Souri\thanks{The first two authors have contributed equally to this work.} \textsuperscript{1}, Pirazh  Khorramshahi\footnotemark[1] \textsuperscript{1}, Chun Pong Lau\textsuperscript{1}, Micah Goldblum\textsuperscript{2}, Rama Chellappa\textsuperscript{1}\\
\textsuperscript{1}Johns Hopkins University, \textsuperscript{2}University of Maryland, College Park \\
{\tt\small \{hsouri1, pkhorra1, clau13, rchella4\}@jhu.edu, goldblum@umd.edu}

}

\maketitle

\begin{abstract}

The adversarial attack literature contains a myriad of algorithms for crafting perturbations which yield pathological behavior in neural networks. In many cases, multiple algorithms target the same tasks and even enforce the same constraints. In this work, we show that different attack algorithms produce adversarial examples which are distinct not only in their effectiveness but also in how they qualitatively affect their victims. We begin by demonstrating that one can determine the attack algorithm that crafted an adversarial example. Then, we leverage recent advances in parameter-space saliency maps to show, both visually and quantitatively, that adversarial attack algorithms differ in which parts of the network and image they target. Our findings suggest that prospective adversarial attacks should be compared not only via their success rates at fooling models but also via deeper downstream effects they have on victims.


\end{abstract}

\section{Introduction}
\label{sec:intro}

\begin{figure}[t]
    \centering
    \subfloat[]{\includegraphics[width=0.4\textwidth]{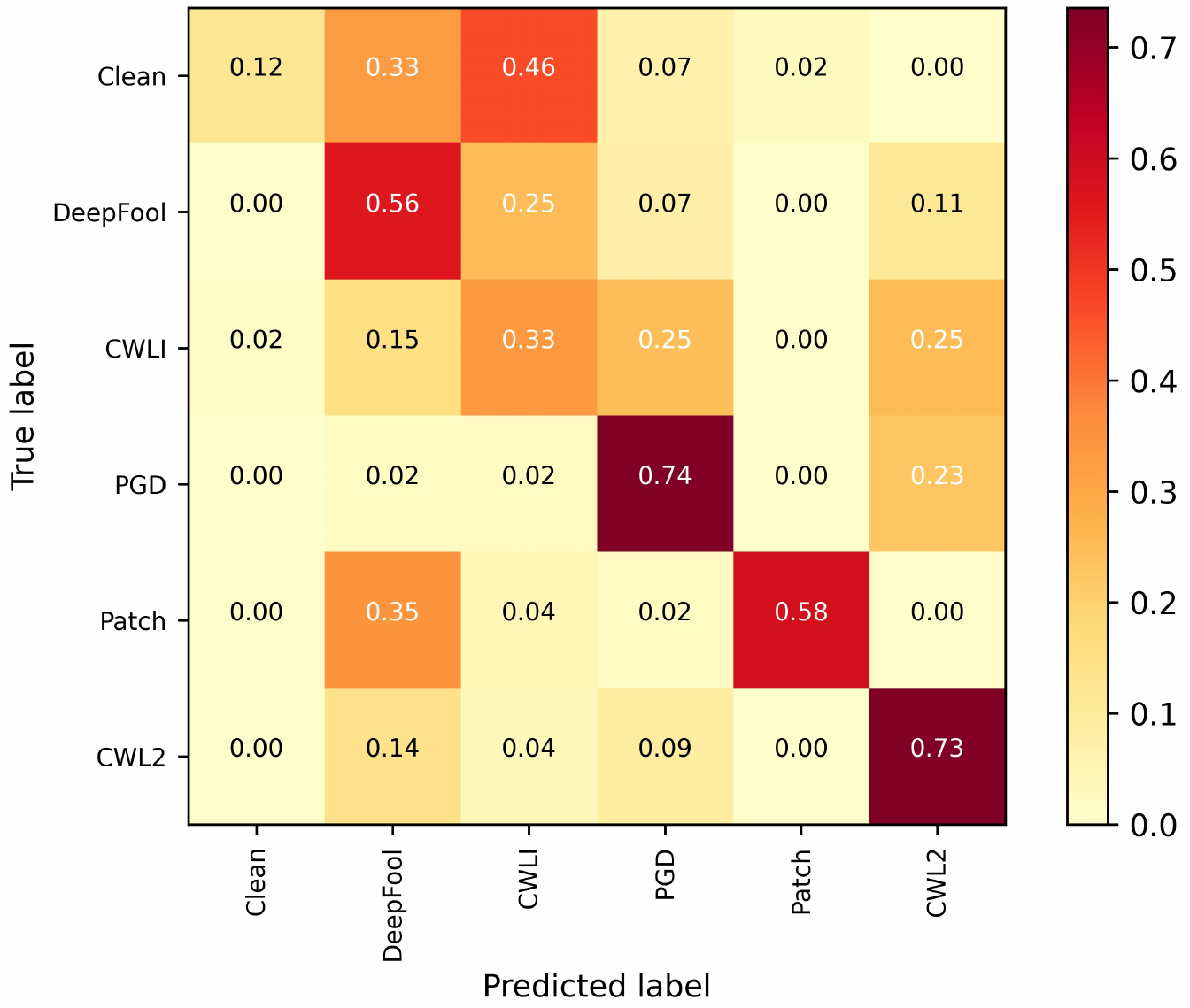}
    \label{fig:IM_confusionmatrix}}\\\vspace{-0.3cm}
    \subfloat[]{\includegraphics[width=0.4\textwidth]{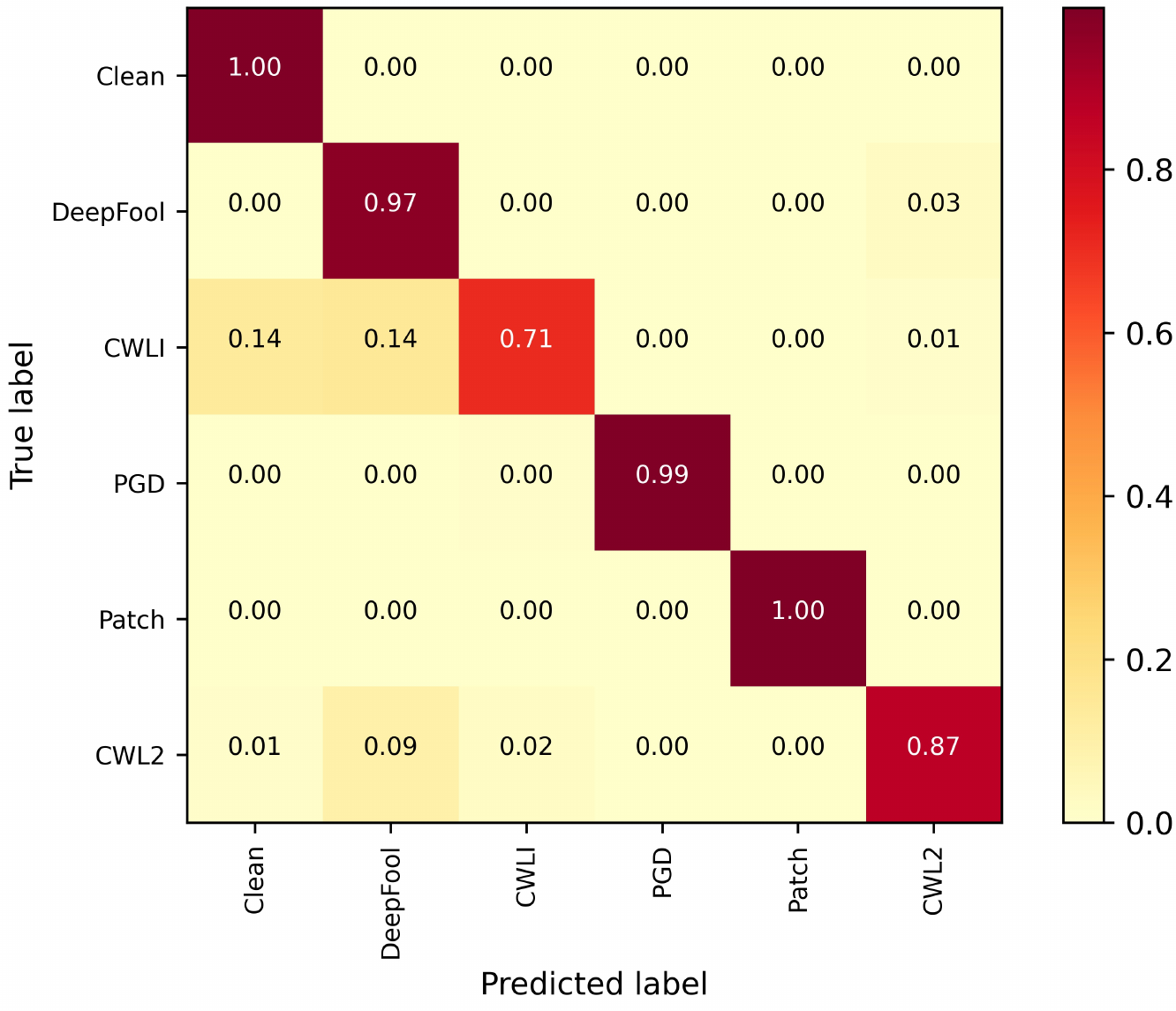}
    \label{fig:RES_confusionmatrix}}
    \caption{(a) Confusion matrix of a ResNet18 network trained on adversarial images for the task of attack type classification. Accuracy of this network is $57.55\%$. (b) Confusion matrix of another ResNet18 network trained on the concatenation of adversarial images with their respective perturbations for the similar task of attack type classification. This network yields the accuracy of $94.23\%$.}
    \label{fig:GT_res_confusionmatrix}
\end{figure}

    

In recent years, neural networks and in particular Deep Convolutional Neural Networks (DCNN) have revolutionized the field of computer vision by outperforming almost all of the traditionally designed methods by a large margin across a variety of tasks including Image Recognition \cite{krizhevsky2017imagenet, he2016deep}, Object Detection \cite{girshick2014rich,liu2016ssd,redmon2016you}, Object Tracking \cite{wang2015visual}, Image Captioning \cite{vinyals2015show}, Face Recognition \cite{ranjan2017hyperface}, Human Pose Estimation \cite{newell2016stacked} and Action Recognition \cite{simonyan2014two}. Despite their performance superiority, DCNNs are easily fooled by maliciously crafted adversarial examples \cite{szegedy2013intriguing}. 

To compromise DCNNs, a number of attacks have been proposed which generate adversarial images that are visually indistinguishable from clean images. Among these attacks are Projected Gradient Descent (PGD) \cite{madry2017towards}, DeepFool (DF) \cite{moosavi2016deepfool}, Carlini \& Wagner (CW) \cite{carlini2017towards}, and Adversarial Patch \cite{brown2017adversarial,karmon2018lavan}. The emergence of adversarial attacks has motivated the development of numerous defense methods for protecting DCNNs against adversaries. Countermeasures against adversaries can be categorized into \textit{Robust Optimization}\cite{lin2020dual}, \textit{Gradient Masking}\cite{papernot2016distillation}, \textit{Adversarial Training} \cite{madry2017towards}, and \textit{Adversarial Example Detection} \cite{xu2019adversarial}. 
Apart from designing attacks and their respective defense methods, we are interested in understanding how different attacks impact their victims in distinct ways and whether we can distinguish the type of an attack based on its resulting adversarial examples. The most straightforward signature of an attack resides in the perturbation it creates on top of the clean sample. However, due to imperceptibility constraints, perturbations are typically much smaller than the original signal. Therefore, it is challenging to recognize and isolate the exact perturbation from the adversarial example alone. 

To investigate whether adversarial attack algorithms can be distinguished via their perturbations and to further understand how unique the signatures in the resulting adversarial examples are, we conduct the following two experiments. We create five adversarial versions of the CIFAR-10 dataset \cite{krizhevsky2009learning} using popular adversarial attacks along with multiple classification models for generating perturbations, multiple attack constraints, and multiple attack hyper-parameters. Then, in the first experiment, we train a classifier to classify images into their respective attack types or the clean class in case they are sampled from the clean training data. In the second experiment, we instead input the concatenation of an adversarial image with its respective perturbation. Figures \ref{fig:IM_confusionmatrix} and \ref{fig:RES_confusionmatrix} show the confusion matrices of attack classifiers from each of these two experiments. From Figure \ref{fig:IM_confusionmatrix}, we observe that the network trained only on the adversarial examples struggles to classify the adversarial images correctly, and in particular, the network fails to distinguish clean samples. On the contrary, Figure \ref{fig:RES_confusionmatrix} demonstrates the high performance of the classifier over the concatenation of perturbations with their respective adversarial examples.  As the network in the second experiment, can recognize attacks much more effectively, \emph{i.e} $94.23\%$ compared to $57.55\%$ recognition accuracy, we conclude that the signature of each attack is unique and encoded in its perturbations.

We note that adversarial perturbations, used in this second experiment, are not readily available for practitioners due to the fact that once an attack occurs, only the corresponding adversarial image is accessible. Therefore, we develop a pipeline for isolating the perturbation from adversarial images based on image residual learning, which has been shown to be an effective tool in capturing small-scale yet important details \cite{khorramshahi2020devil}. Residual images enable us to identify attack algorithms from the adversarial example alone effectively.

The observation that adversarial attack algorithms generate distinct examples motivates us to explore how these attacks manifest their differences in the downstream behavior of victim models. Specifically, we adopt the parameter-space saliency maps method \cite{levin2021models} to understand which parts of the victim model are targeted by a particular attack model.  We find that attack algorithms exhibit not only superficially distinguishable image patterns but also influence their victims in unique qualitative fashions.

The rest of the paper is organized as follows. In Section \ref{sec:related_work}, we review recent related works in the area of adversarial attacks, defenses and saliency maps. Next, we discuss the datasets and adversarial attack algorithms which are used in this work in section \ref{sec:exp_setup}. Section \ref{sec:method} describes our proposed method for the estimation of adversarial perturbations. Throughout section \ref{sec:saliency}, we investigate how different attack methods manipulate the behaviour of neural networks. Finally we conclude the paper in section \ref{conclusion}.
\section{Related Work}
\label{sec:related_work}

\subsection{Adversarial Attacks}
Adversarial attacks perturb inputs to models in order to degrade their performance \cite{szegedy2013intriguing}.  Typical attacks employ optimizers to maximize the model’s loss or a surrogate loss with respect to the input constrained to a set which ensures that the resulting adversarial example is hard to be distinguished from its original counterpart. One famous attack, projected gradient descent (PGD), involves an alternating procedure of signed gradient ascent steps on cross-entropy loss and projection onto the constraint set, typically an $\ell_{\infty}$ ball surrounding the base image \cite{madry2017towards}. \cite{carlini2017towards} explores numerous alternative surrogate objective functions and finds that surrogates can improve the effectiveness of adversaries over directly optimizing cross-entropy loss. Adversarial attacks have been proposed both to probe the robustness of neural networks \cite{xu2018structured, ozbulak2021investigating} and to compromise their security \cite{goldblum2020adversarial, cherepanova2021lowkey}.
 
In response to this looming threat, a number of defenses have been proposed \cite{madry2017towards, zhang2019theoretically, goldblum2020adversarially}.  Another line of defensive works involves detecting adversarial examples to alert critical systems \cite{metzen2017detecting, feinman2017detecting, grosse2017statistical}.  While this direction may appear similar to our own on the surface, our methods do not focus on distinguishing adversarial and clean examples but instead seek to distinguish adversarial examples generated via different algorithms from each other. Moreover, our methods are not intended as defensive tools but instead provide exploratory observations regarding the nature of adversarial examples.
 
\subsection{Saliency Maps}
A major component of our analysis relies on \emph{saliency maps}.  Typical saliency maps identify the extent to which each input feature (\emph{e.g.} pixel) of a particular sample influences a model’s output \cite{simonyan2013deep, selvaraju2017grad}.  One recent work instead identifies the extent to which each model parameter is responsible for model misbehaviors \cite{levin2021models}.  This tool allows us to examine precisely which components in a neural network are affected by an adversarial example and also how these components interact with the input. We find that, in fact, different adversarial attacks influence different parts of the network, even when we average over thousands of samples.

\section{Experimental Setup}
\label{sec:exp_setup}
\begin{figure*}[t!]
    \centering
    \subfloat[][DeepFool]{\includegraphics[width=.12\textwidth]{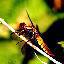}}~
    \subfloat[][CW$L_2$]{\includegraphics[width=.12\textwidth]{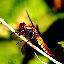}}~
    \subfloat[][CW$L_\infty$]{\includegraphics[width=.12\textwidth]{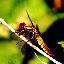}}~
    \subfloat[][PGD]{\includegraphics[width=.12\textwidth]{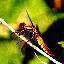}}~
    \subfloat[][Patch]{\includegraphics[width=.12\textwidth]{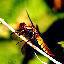}}\\\vspace{-.3cm}
    \subfloat[][]{\includegraphics[width=.12\textwidth]{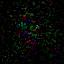}}~
    \subfloat[][]{\includegraphics[width=.12\textwidth]{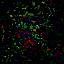}}~
    \subfloat[][]{\includegraphics[width=.12\textwidth]{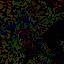}}~
    \subfloat[][]{\includegraphics[width=.12\textwidth]{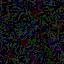}}~
    \subfloat[][]{\includegraphics[width=.12\textwidth]{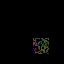}}\\
    \caption{Adversarial Samples (first row) and their respective relatively sparse perturbations (second row), generated based on a ResNext50 network trained on the Tiny ImageNet dataset \cite{wu2017tiny}. Note for the purpose of visualization, second row images have been multiplied by a factor to become visible.}
    \label{fig:adversarial_samples_vs_perturbation}
\end{figure*}

\subsection{Datasets}
In this study, we consider the CIFAR-10 and Tiny ImageNet \cite{wu2017tiny} datasets. Tiny ImageNet is a smaller subset of the ImageNet benchmark \cite{deng2009imagenet} with $200$ classes. Compared to CIFAR-10, Tiny ImageNet has a higher diversity and larger dimensionality, \emph{i.e.} images are of the size $64\times64$. 

\subsection{Adversarial Attacks}
\label{subsec:adversarial_attacks}
To generate adversarial perturbations required for training and testing in our experiments, we considered the following attacks:
\begin{itemize}
    \item \textbf{PGD}\cite{madry2017towards} produces adversarial perturbation, \emph{i.e} $\delta$, in an iterative fashion in multiple steps. At each iteration, a step is taken in the signed gradient direction to maximize the cross-entropy loss of the victim model. After every such step, the attacker projects $\delta$ onto the $\ell_\infty$ ball in order to ensure that the generated adversarial sample $I_{adv}$ remains within the constraint space surrounding clean image $I_c$.
    \item \textbf{DeepFool}\cite{moosavi2016deepfool} attempts to iteratively find a route for $I_c$ to pass the decision boundary drawn by the an image classification model. Here, $\delta$ can be computed as the summation of perturbations (normal vector of the linearized decision boundary) found at each iteration.
    \item \textbf{CW} \cite{carlini2017towards} is a computationally expensive iterative approach to generating adversarial samples which instead opts for a surrogate objective formed by subtracting the highest incorrect logit from the logit corresponding to the correct class.  In this work, we consider $L_2$ and $L_\infty$ norm versions of the CW attack. 
    \item \textbf{Adversarial Patch}\cite{karmon2018lavan}
    selects a random location within a clean sample to be overlaid with a patch with desired height and width. In addition, we ensure that the crafted patch remains within the $\ell_\infty$-ball surrounding the corresponding clean image in the patch location. 
\end{itemize}

\section{Adversarial Perturbation Estimation}
\label{sec:method}

\begin{figure*}[t]
    \centering
    \includegraphics[width=.7\textwidth]{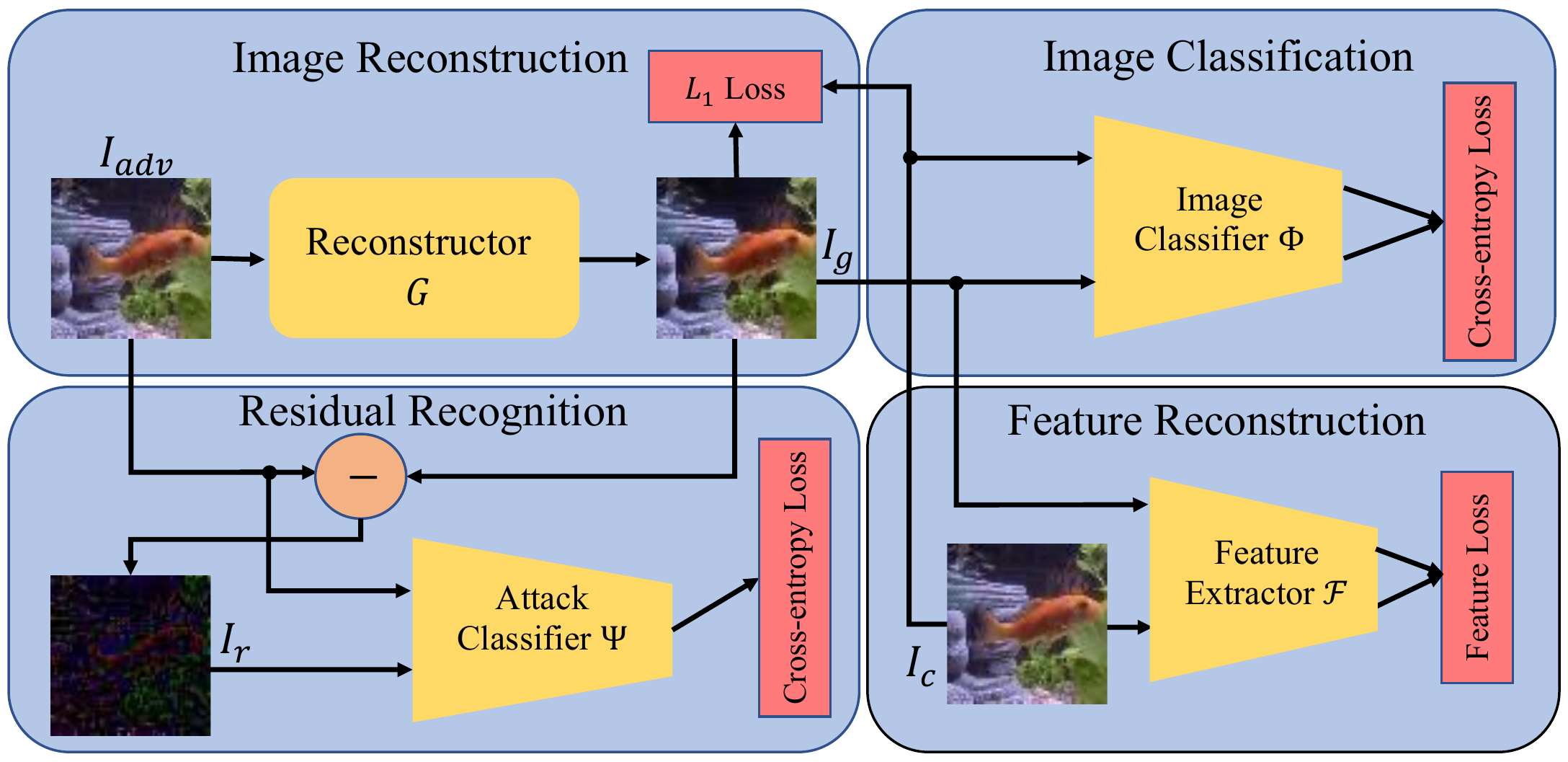}
    \caption{Adversarial Perturbation Recovery via Reverse Engineering of Deceptions via Residual Learning (REDRL) Pipeline. Network $G$ reconstructs the input adversarial image $I_{adv}$ and generates $I_g$ to lie close to the corresponding clean image $I_c$ in $L_1$ distance. $I_g$ should also result in a high classification score passing through the frozen image classifier $\Phi$ trained on $I_c$. Additionally, $G$ produces $I_g$ so that it's extracted features by general feature extractor $\mathcal{F}$, lie close to that of $I_c$. To recognize an attack type, generated image $I_g$ is subtracted from the input $I_{adv}$ to obtain the residual image $I_r$, an estimate of adversarial perturbation. Subsequently, $\Psi$ model classifies concatenation of $I_r$ with its respective $I_{adv}$ into its corresponding class.} 
    \label{fig:Method}
\end{figure*}

Adversarial perturbations are relatively sparse and generally carry much lower energy compared to the original image's content as seen in Figure \ref{fig:adversarial_samples_vs_perturbation}. In addition, Figure \ref{fig:GT_res_confusionmatrix} demonstrates the advantage of including perturbations over adversarial images alone when classifying attack algorithms. However, in reality, networks do not have access to the adversarial perturbations. Therefore, we propose a method for estimating adversarial perturbations in which the roles of signal and noise are switched. In other words, rather than treating clean images as the signal of interest and adversarial perturbation as noise, we switch our focus to first recovering perturbations and then using them to classify attack algorithms.

In an effort to estimate adversarial perturbations, recognize the presence of attacks, and determine their parent algorithms, we develop Reverse Engineering of Deceptions via Residual Learning (REDRL) where a set of constraints are introduced to help the Reconstruction network $G$ in recovering the clean image components from adversarial images so that adversarial perturbations can be isolated. The pipeline of REDRL is outlined in Figure \ref{fig:Method} which consists of four main components, namely \textit{image reconstruction}, \textit{image classification}, \textit{feature reconstruction}, and \textit{residual recognition}. During training, REDRL takes an image as the input $I_{adv} = I_c + \delta$ where $I_c$ and $\delta$ denote a clean image and adversarial perturbation, respectively. REDRL tries to salvage the content in the input image $I_{adv}$ so that the generated image $I_g$ lies close to the manifold of clean images. To ensure this property, REDRL constrains the generated image with several objectives, which we now enumerate. 

\subsection{Image Reconstruction}
\label{subsec:image_recon}
First, $I_g$ should lie close to $I_c$ in pixel space. Different distance metrics can be used for this purpose; however, we choose the $L_1$ distance similar to the work of Pix2Pix \cite{8100115}. For the architecture of reconstructor $G$, we leverage prior work in super-resolution \cite{ledig2017photo}. Therefore, $G$ is composed of a set of identical residual blocks without batch normalization that are preceded and followed by two convolutional layers with Leaky ReLU non-linearities. Note that there is no skip connection in $G$ as we would like to eliminate any direct transportation of adversarial perturbations. In addition, the spatial size of the input image is maintained during the reconstruction process. Our image reconstruction objective encourages $G$ to recover $I_g = G(I_{adv})$ by minimizing the following function: 

\begin{equation}
\mathcal{L}_{R}(G) = \underset{{I_c, \delta}}{\mathbb{E}} \biggl[ {|I_c- G(I_c + \delta)|}_1\biggr]
\label{eq:recon_loss}
\end{equation}

\subsection{Feature Reconstruction}
\label{subsec:feature_recon}
In addition to the previous pixel-space objective, $I_g$ should lie close to $I_c$ in the feature space of a neural network feature extractor so that $I_g$ is semantically similar to $I_c$. Inspired by the work of \cite{Naseer_2020_CVPR}, we employ the feature space of the third convolutional block of an ImageNet-trained VGG16 network \cite{simonyan2014very} $\mathcal{F}$ and $L_2$ distance to minimize the feature distortion between $I_g$ and $I_c$. Subsequently, our feature reconstruction objective encourages $G$ to minimize the following loss function:
\begin{equation}
    \mathcal{L}_{F}(G) = \underset{{I_c, \delta}}{\mathbb{E}} \biggl[ {|\mathcal{F}(I_c) - \mathcal{F}(G(I_c + \delta))|}_2\biggr]
    \label{eq:feature_loss}
\end{equation}

\subsection{Image Classification}
\label{subsec:image_cls}
Moreover, REDRL ensures that its reconstructions do not affect the output of an image classifier $\Phi$, trained on clean images. If $G$ is successful at salvaging the critical components of $I_c$ and modeling its distribution, $\Phi$ should yield similar classification scores on $I_g$ and $I_c$. This objective can be framed in the context of Knowledge Distillation \cite{Chen_2019_ICCV}. To this end, we introduce the following cross-entropy loss function to REDRL when training $G$:
\begin{equation}
    \mathcal{L}_{IC}(G) = \underset{{I_c, \delta}}{\mathbb{E}} \biggl[ -\log(\frac{e^{\Phi_{i}(G(I_c + \delta))}}{\sum_{j=1}^{C}e^{\Phi_{j}(G(I_c + \delta))}})\biggr]
    \label{eq:image_classification_loss}
\end{equation}
In Eq. \ref{eq:image_classification_loss}, $i$, $\Phi_i$, and $C$ are the $\text{argmax} ~ \Phi(I_c)$, $i^{th}$ logit in the output of $\Phi$, and the total number of image classes, respectively. During the training of REDRL, the parameters of $\Phi$ are frozen and the gradients from $\mathcal{L}_{IC}$ are back-propagated through the reconstructor $G$ to ensure that $I_g$ is similar to $I_c$ in the view of an image classifier.

\subsection{Residual Recognition}
\label{subsec:residual_recognition}
Finally, the residual recognition objective subtracts the generated image from the input to obtain the residual image $I_r$. As REDRL seeks to generate an image close to the clean images manifold, $I_r$ is a reasonable estimate of adversarial perturbations. Next, $I_r$ is fed to the attack classification network $\Psi$ to be classified into one of the respective adversarial attack algorithm classes.

The end goal of REDRL is to estimate the adversarial perturbations and recognize their parent algorithms. Therefore, as the network $G$ reconstructs the input $I_{adv}$ to be close to its corresponding clean image $I_c$,  $I_g$ is subtracted from $I_{adv}$ to obtain the residual image, $I_r$, an estimate of the adversarial perturbation. Consequently, much of the remaining information in the residual image is representative of the signature of the attack that can be used in its recognition. The attack recognition model $\Psi$ is trained to classify the concatenation of $I_{adv}$ and $I_r$ into the desired class through the attack classification cross-entropy loss function
\begin{equation}
    \mathcal{L}_{AC}(G) = \underset{{I_c, \delta}}{\mathbb{E}}\biggl[- 
    \log(\frac{e^{\Psi_{i}(I_r, I_c + \delta)}}{\sum_{j=1}^{A}e^{\Psi_{j}(I_r, I_c + \delta)}})
    \biggr]
    \label{eq:attack_classification_loss}
\end{equation}
where $A$, $i$, and $\Psi_j$ are the total number of considered attack types plus one (for clean data), the true class label corresponding to $I_r$, and $j^{th}$ logit of $\Psi(I_r, I_adv)$, respectively. It is noteworthy to mention that the gradient generated from $\mathcal{L}_{AC}$ also back-propagates through the network $G$ so that the obtained residuals carry information needed for attack recognition.

\subsection{End-To-End Training}
The four stages of REDRL are trained simultaneously in an end-to-end fashion for the purpose of adversarial perturbations estimation and attack algorithm recognition. Altogether, the optimization objective for the end-to-end training of REDRL can be calculated as follows:
\begin{equation}
    \mathcal{L}_{total} = \min_{G} \biggl[\mathcal{L}_{AC}(G) + \lambda_1 \mathcal{L}_{R}(G) + \lambda_2 \mathcal{L}_{F}(G) + \lambda_3 \mathcal{L}_{IC}(G)\biggr]
    \label{eq:loss_total}
\end{equation}
In Eq. \ref{eq:loss_total}, weights $\lambda_1$, $\lambda_2$, and $\lambda_3$ are empirically set to $0.01$, $0.1$, and $1.0$, respectively. 

\begin{table}
    \centering
    \caption{Image classification accuracy ($\%$) of different image classification models $\Phi$ over test set of \textbf{clean} samples for CIFAR-10 and Tiny ImageNet datasets.\\}
    \resizebox{\columnwidth}{!}{%
    \begin{tabular}{c|c|c|c|c}
         \cline{1-5}
         \multicolumn{1}{|c|}{\multirow{2}{*}{Dataset}} & 
         \multicolumn{4}{|c|}{Model Architecture} \\
         \cline{2-5}
         \multicolumn{1}{|c}{} &
         \multicolumn{1}{|c|}{ResNet50} & \multicolumn{1}{|c|}{ResNext50} & \multicolumn{1}{|c|}{DenseNet121} & \multicolumn{1}{|c|}{VGG19} \\
         \hline
         \hline
         \multicolumn{1}{|c|}{CIFAR-10} & 94.02 & 94.76 & 94.48 & \multicolumn{1}{|c|}{95.73} \\
         \cline{1-5}
         \multicolumn{1}{|c|}{Tiny ImageNet}& 65.18 & 68.05 & 65.03 & \multicolumn{1}{|c|}{64.89} \\
         \hline
    \end{tabular}
    }
    \label{tab:image_classifier_details}
\end{table}

\subsection{Implementation Details}
In this section, we discuss the implementation details of the proposed REDRL and its different modules. 

For both CIFAR-10 and Tiny ImageNet, ResNet50, ResNext50, DenseNet121 and VGG19 are used for the choices of architecture of $\Phi$. Table \ref{tab:image_classifier_details} reports the accuracy of different $\Phi$ models on the clean test data after being optimized for $50$ epochs with Stochastic Gradient Descent with Nestrov Momentum \cite{sutskever2013importance}, initial learning rate of $0.01$ and learning rate decay factor of $10$ on ${25}^{th}$ and ${40}^{th}$ epochs. Using the trained image classifiers, and the five attack algorithms discussed in subsection \ref{subsec:adversarial_attacks}, we create attacked versions of CIFAR-10 and Tiny ImageNet datasets. Table \ref{tab:attack_configs} summarizes the configurations and hyper-parameters we used for each attack to create the respective adversarial datasets. 

\begin{table}[!htb]
    \centering
    \caption{Adversarial Attacks Settings}
     \resizebox{\columnwidth}{!}{%
    \begin{tabular}{|c|c|}
    \hline
    \multicolumn{1}{|c|}{Attack Type} & \multicolumn{1}{|c|}{Configuration} \\
    \hline
    \hline
    \multicolumn{1}{|c|}{DeepFool} & Steps: $50$ \\
    \hline
    \hline
    \multicolumn{1}{|c}{\multirow{2}{*}{PGD}} & \multicolumn{1}{|c|}{$\epsilon \in \{4,8,16\}$} \\
    \cline{2-2}
    \multicolumn{1}{|c}{} & \multicolumn{1}{|c|}{$\alpha: 0.01$, Steps: $100$} \\
    \hline
    \hline
    \multicolumn{1}{|c}{\multirow{2}{*}{CW$L_2$}} &  \multicolumn{1}{|c|}{Steps: $1000$, $c \in \{100, 1000\}$} \\
    \cline{2-2}
    \multicolumn{1}{|c}{} & \multicolumn{1}{|c|}{Learning Rate: $0.01$, $\kappa : 0$} \\
    \hline
    \hline
    \multicolumn{1}{|c}{\multirow{2}{*}{CW$L_\infty$}} &  \multicolumn{1}{|c|}{Steps: $100$, $\epsilon \in \{4,8,16\}$} \\
    \cline{2-2}
    \multicolumn{1}{|c}{} & \multicolumn{1}{|c|}{Learning Rate: $0.005$, c: $5$} \\
    \hline
    \hline
    \multicolumn{1}{|c}{\multirow{2}{*}{Adversarial Patch}} & \multicolumn{1}{|c|}{Steps: $100$, $\epsilon \in \{4,8,16\}$} \\
    \cline{2-2}
    \multicolumn{1}{|c}{} & \multicolumn{1}{|c|}{Patch Size $\in \{4\times4, 8\times8, 16 \times 16\}$} \\
    \hline
    \end{tabular}
    }
    \label{tab:attack_configs}
\end{table}


As mentioned in subsection \ref{subsec:image_recon}, for the reconstructor network $G$, we use a cascade of residual blocks without batch normalization layer and skip connections. Moreover, for the residual recognition network $\Psi$, we consider ResNet18 model to recognize different attack types based on the concatenation of computed residual $I_r$ with the respective adversarial image $I_{adv}$. To prevent $\Psi$ from over-fitting the training data and latching onto undesired patterns in the residuals, we use the label smoothing technique proposed in \cite{szegedy2016rethinking}. Lastly, the REDRL pipeline is trained end-to-end, with a learning rate warm-up technique \cite{fan2019spherereid} in the first $10$ epochs that has been shown to contribute to the improved recognition performance of DCNNs. In total REDRL is trained for $100$ epochs with the Adam optimizer \cite{kingma2014adam} and learning rates of $0.00035$ and $0.00017$ for CIFAR-10 and Tiny ImageNet datasets, respectively.

\subsection{Experimental Evaluation}
In section \ref{sec:intro}, we performed two experiments to show whether it is possible to distinguish adversarial attack types based on their signatures. In addition, Figure \ref{fig:GT_res_confusionmatrix} demonstrates the degree to which this distinguishability varies from adversarial images $I_{adv}$ to adversarial perturbations $\delta$. Consequently, in the absence of true perturbations and with the goal of having high attack recognition capability, we propose REDRL to estimate adversarial perturbations, \emph{i.e.} $I_r$. Here we present the evaluation results of REDRL and highlight how the recognition performance gap from adversarial images and adversarial perturbations as seen in Figure \ref{fig:GT_res_confusionmatrix}, can be compensated with the help of residual images.  

\begin{table}[t!]
    \centering
    \caption{Performance comparison (\%) of attack algorithm type recognition based on adversarial images $I_{adv}$, concatenation of ground-truth adversarial perturbations $\delta$ with $I_{adv}$, and concatenation of estimated residuals $I_r$ with $I_{adv}$, \emph{i.e.}, REDRL.\\}
    \resizebox{\columnwidth}{!}{
    \begin{tabular}{c|c|c|c|c|c|c}
         \cline{2-7}
         \multicolumn{1}{c|}{} & \multicolumn{6}{|c|}{Dataset} \\
         \cline{1-7}
         \multicolumn{1}{|c|}{\multirow{3}{*}{Class}} & \multicolumn{3}{|c||}{CIFAR-10} & \multicolumn{3}{|c|}{Tiny ImageNet}\\
         \cline{2-7}
         \multicolumn{1}{|c|}{} & \multicolumn{3}{|c||}{Input to $\Psi $} & \multicolumn{3}{|c|}{Input to $\Psi$}\\
         \cline{2-7}
         \multicolumn{1}{|c|}{} & \multicolumn{1}{|c|}{$I_{adv}$} & \multicolumn{1}{|c|}{$\delta, I_{adv}$} & \multicolumn{1}{|c||}{$I_r, I_{adv}$} & \multicolumn{1}{|c|}{$I_{adv}$} & \multicolumn{1}{|c|}{$\delta, I_{adv}$} & \multicolumn{1}{|c|}{$I_r, I_{adv}$} 
         \\
         \cline{1-7}
         \multicolumn{1}{|c|}{Clean} & \multicolumn{1}{|c|}{$12.0$} & \multicolumn{1}{|c|}{$100$} & \multicolumn{1}{|c||}{$100$} & \multicolumn{1}{|c|}{$62.5$} & \multicolumn{1}{|c|}{$99.9$} & \multicolumn{1}{|c|}{$99.7$}\\
         \cline{1-7}
         \multicolumn{1}{|c|}{PGD} & \multicolumn{1}{|c|}{$73.5$} & \multicolumn{1}{|c|}{$99.9$} & \multicolumn{1}{|c||}{$99.9$} & \multicolumn{1}{|c|}{$88.7$} & \multicolumn{1}{|c|}{$99.7$} & \multicolumn{1}{|c|}{$99.9$} \\
         \cline{1-7}
         \multicolumn{1}{|c|}{DeepFool} & \multicolumn{1}{|c|}{$56.2$} & \multicolumn{1}{|c|}{$99.9$} & \multicolumn{1}{|c||}{$97.4$} & \multicolumn{1}{|c|}{$53.2$} & \multicolumn{1}{|c|}{$64.0$} & \multicolumn{1}{|c|}{$75.3$} \\
         \cline{1-7}
         \multicolumn{1}{|c|}{CW$L_2$} & \multicolumn{1}{|c|}{$73.4$} & \multicolumn{1}{|c|}{$98.6$} & \multicolumn{1}{|c||}{$96.6$} & \multicolumn{1}{|c|}{$28.0$} & \multicolumn{1}{|c|}{$96.4$} & \multicolumn{1}{|c|}{$66.3$} \\
         \cline{1-7}
         \multicolumn{1}{|c|}{CW$L_\infty$} & \multicolumn{1}{|c|}{$33.4$} & \multicolumn{1}{|c|}{$71.6$} & \multicolumn{1}{|c||}{$74.1$} & \multicolumn{1}{|c|}{$24.2$} & \multicolumn{1}{|c|}{$92.7$} & \multicolumn{1}{|c|}{$57.7$} \\
         \cline{1-7}
         \multicolumn{1}{|c|}{Patch} & \multicolumn{1}{|c|}{$58.4$} & \multicolumn{1}{|c|}{$99.9$} & \multicolumn{1}{|c||}{$99.9$} & \multicolumn{1}{|c|}{$73.8$} & \multicolumn{1}{|c|}{$99.9$} & \multicolumn{1}{|c|}{$99.6$} \\
          \hline
         \hline
         \multicolumn{1}{|c|}{Total} & \multicolumn{1}{|c|}{$57.5$} & \multicolumn{1}{|c|}{$94.2$} & \multicolumn{1}{|c||}{$94.2$} & \multicolumn{1}{|c|}{$59.4$} & \multicolumn{1}{|c|}{$95.7$} & \multicolumn{1}{|c|}{$85.5$}\\
         \cline{1-7}
    \end{tabular}
    }
    \label{tab:REDRL}
\end{table}

\begin{table*}[t!]
    \centering
    \caption{Comparison of image classification accuracy (\%) of different classification models on images before and after being applied to the reconstructor network $G$. Note that $I$ and $G(I)$ represent input images to and reconstructed images from $G$ respectively.\\}
    \resizebox{1.5\columnwidth}{!}{
    \begin{tabular}{c|c|c|c|c|c|c|c|c|c|c|c|c|}
         \cline{1-13}
         \multicolumn{1}{|c|}{\multirow{3}{*}{Network}} & \multicolumn{12}{|c|}{CIFAR-10}\\
         \cline{2-13}
         \multicolumn{1}{|c|}{} & 
         \multicolumn{2}{|c|}{Clean} & \multicolumn{2}{|c|}{DeepFool} & \multicolumn{2}{|c|}{PGD} & \multicolumn{2}{|c|}{CW$L_2$} & \multicolumn{2}{|c|}{CW$L_\infty$} & \multicolumn{2}{|c|}{Patch} \\
         \cline{2-13}
         \multicolumn{1}{|c|}{} & 
         \multicolumn{1}{|c|}{$I$} &
         \multicolumn{1}{|c|}{$G(I)$} &
         \multicolumn{1}{|c|}{$I$} &
         \multicolumn{1}{|c|}{$G(I)$} &
         \multicolumn{1}{|c|}{$I$} &
         \multicolumn{1}{|c|}{$G(I)$} &
         \multicolumn{1}{|c|}{$I$} &
         \multicolumn{1}{|c|}{$G(I)$} &
         \multicolumn{1}{|c|}{$I$} &
         \multicolumn{1}{|c|}{$G(I)$} &
         \multicolumn{1}{|c|}{$I$} &
         \multicolumn{1}{|c|}{$G(I)$}\\
         \cline{1-13}
         \multicolumn{1}{|c|}{ResNet50} & 
         \multicolumn{1}{|c|}{$94.0$} &
         \multicolumn{1}{|c|}{$94.0$} &
         \multicolumn{1}{|c|}{$7.4$} &
         \multicolumn{1}{|c|}{$87.6$} &
         \multicolumn{1}{|c|}{$2.1$} &
         \multicolumn{1}{|c|}{$90.9$} &
         \multicolumn{1}{|c|}{$0.1$} &
         \multicolumn{1}{|c|}{$88.2$} &
         \multicolumn{1}{|c|}{$13.5$} &
         \multicolumn{1}{|c|}{$85.0$} &
         \multicolumn{1}{|c|}{$79.0$} &
         \multicolumn{1}{|c|}{$93.3$}\\
         \cline{1-13}
         \multicolumn{1}{|c|}{ResNext50} & 
         \multicolumn{1}{|c|}{$94.7$} &
         \multicolumn{1}{|c|}{$94.7$} &
         \multicolumn{1}{|c|}{$8.0$} &
         \multicolumn{1}{|c|}{$87.0$} &
         \multicolumn{1}{|c|}{$2.41$} &
         \multicolumn{1}{|c|}{$91.3$} &
         \multicolumn{1}{|c|}{$0.0$} &
         \multicolumn{1}{|c|}{$85.5$} &
         \multicolumn{1}{|c|}{$12.2$} &
         \multicolumn{1}{|c|}{$85.3$} &
         \multicolumn{1}{|c|}{$80.3$} &
         \multicolumn{1}{|c|}{$94.0$}\\
         \cline{1-13}
         \multicolumn{1}{|c|}{DenseNet121} & 
         \multicolumn{1}{|c|}{$94.4$} &
         \multicolumn{1}{|c|}{$94.4$} &
         \multicolumn{1}{|c|}{$7.9$} &
         \multicolumn{1}{|c|}{$86.3$} &
         \multicolumn{1}{|c|}{$2.1$} &
         \multicolumn{1}{|c|}{$90.6$} &
         \multicolumn{1}{|c|}{$0.6$} &
         \multicolumn{1}{|c|}{$80.9$} &
         \multicolumn{1}{|c|}{$13.0$} &
         \multicolumn{1}{|c|}{$84.1$} &
         \multicolumn{1}{|c|}{$78.6$} &
         \multicolumn{1}{|c|}{$93.7$}\\
         \cline{1-13}
         \multicolumn{1}{|c|}{VGG19} & 
         \multicolumn{1}{|c|}{$95.7$} &
         \multicolumn{1}{|c|}{$95.7$} &
         \multicolumn{1}{|c|}{$8.4$} &
         \multicolumn{1}{|c|}{$87.1$} &
         \multicolumn{1}{|c|}{$2.0$} &
         \multicolumn{1}{|c|}{$93.1$} &
         \multicolumn{1}{|c|}{$0.2$} &
         \multicolumn{1}{|c|}{$84.3$} &
         \multicolumn{1}{|c|}{$5.4$} &
         \multicolumn{1}{|c|}{$86.2$} &
         \multicolumn{1}{|c|}{$75.8$} &
         \multicolumn{1}{|c|}{$94.9$}\\
         \hline
    \end{tabular}
    }
    \vspace{0.5cm}
    
    \resizebox{1.5\columnwidth}{!}{
    \begin{tabular}{c|c|c|c|c|c|c|c|c|c|c|c|c|}
         \cline{1-13}
         \multicolumn{1}{|c|}{\multirow{3}{*}{Network}} & \multicolumn{12}{|c|}{Tiny ImageNet}\\
         \cline{2-13}
         \multicolumn{1}{|c|}{} & 
         \multicolumn{2}{|c|}{Clean} & \multicolumn{2}{|c|}{DeepFool} & \multicolumn{2}{|c|}{PGD} & \multicolumn{2}{|c|}{CW$L_2$} & \multicolumn{2}{|c|}{CW$L_\infty$} & \multicolumn{2}{|c|}{Patch} \\
         \cline{2-13}
         \multicolumn{1}{|c|}{} &
         \multicolumn{1}{|c|}{$I$} &
         \multicolumn{1}{|c|}{$G(I)$} &
         \multicolumn{1}{|c|}{$I$} &
         \multicolumn{1}{|c|}{$G(I)$} &
         \multicolumn{1}{|c|}{$I$} &
         \multicolumn{1}{|c|}{$G(I)$} &
         \multicolumn{1}{|c|}{$I$} &
         \multicolumn{1}{|c|}{$G(I)$} &
         \multicolumn{1}{|c|}{$I$} &
         \multicolumn{1}{|c|}{$G(I)$} &
         \multicolumn{1}{|c|}{$I$} &
         \multicolumn{1}{|c|}{$G(I)$}\\
         \cline{1-13}
         \multicolumn{1}{|c|}{ResNet50} & 
         \multicolumn{1}{|c|}{$65.1$} &
         \multicolumn{1}{|c|}{$65.1$} &
         \multicolumn{1}{|c|}{$11.5$} &
         \multicolumn{1}{|c|}{$51.3$} &
         \multicolumn{1}{|c|}{$0.3$} &
         \multicolumn{1}{|c|}{$37.8$} &
         \multicolumn{1}{|c|}{$0.1$} &
         \multicolumn{1}{|c|}{$39.5$} &
         \multicolumn{1}{|c|}{$2.7$} &
         \multicolumn{1}{|c|}{$40.9$} &
         \multicolumn{1}{|c|}{$48.5$} &
         \multicolumn{1}{|c|}{$61.0$}\\
         \cline{1-13}
         \multicolumn{1}{|c|}{ResNext50} & 
         \multicolumn{1}{|c|}{$68.0$} &
         \multicolumn{1}{|c|}{$68.0$} &
         \multicolumn{1}{|c|}{$11.2$} &
         \multicolumn{1}{|c|}{$53.9$} &
         \multicolumn{1}{|c|}{$0.4$} &
         \multicolumn{1}{|c|}{$36.6$} &
         \multicolumn{1}{|c|}{$0.1$} &
         \multicolumn{1}{|c|}{$36.5$} &
         \multicolumn{1}{|c|}{$3.1$} &
         \multicolumn{1}{|c|}{$43.5$} &
         \multicolumn{1}{|c|}{$51.6$} &
         \multicolumn{1}{|c|}{$63.8$}\\
         \cline{1-13}
         \multicolumn{1}{|c|}{DenseNet121} &
         \multicolumn{1}{|c|}{$65.0$} &
         \multicolumn{1}{|c|}{$64.9$} &
         \multicolumn{1}{|c|}{$11.8$} &
         \multicolumn{1}{|c|}{$53.2$} &
         \multicolumn{1}{|c|}{$0.1$} &
         \multicolumn{1}{|c|}{$27.3$} &
         \multicolumn{1}{|c|}{$0.1$} &
         \multicolumn{1}{|c|}{$31.6$} &
         \multicolumn{1}{|c|}{$3.2$} &
         \multicolumn{1}{|c|}{$39.0$} &
         \multicolumn{1}{|c|}{$48.8$} &
         \multicolumn{1}{|c|}{$60.4$}\\
         \cline{1-13}
         \multicolumn{1}{|c|}{VGG19} & 
         \multicolumn{1}{|c|}{$64.8$} &
         \multicolumn{1}{|c|}{$64.8$} &
         \multicolumn{1}{|c|}{$11.1$} &
         \multicolumn{1}{|c|}{$45.9$} &
         \multicolumn{1}{|c|}{$0.2$} &
         \multicolumn{1}{|c|}{$26.1$} &
         \multicolumn{1}{|c|}{$0.0$} &
         \multicolumn{1}{|c|}{$31.5$} &
         \multicolumn{1}{|c|}{$1.6$} &
         \multicolumn{1}{|c|}{$34.1$} &
         \multicolumn{1}{|c|}{$45.7$} &
         \multicolumn{1}{|c|}{$59.5$}\\
         \hline
    \end{tabular}
    }
    \label{tab:before_after}
\end{table*}

\begin{figure}
    \centering
    \subfloat[]{\includegraphics[width=0.15\textwidth]{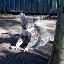}}\hspace{0.01cm}
    \subfloat[]{\includegraphics[width=0.15\textwidth]{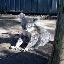}
    }
    \subfloat[]{\includegraphics[width=0.15\textwidth]{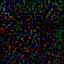}
    }
    \\
    \vspace{-0.3cm}
    \subfloat[]{\includegraphics[width=0.15\textwidth]{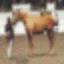}
    }
    \subfloat[]{\includegraphics[width=0.15\textwidth]{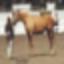}}\hspace{0.01cm}
    \subfloat[]{\includegraphics[width=0.15\textwidth]{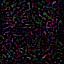}}
    \caption{Sample adversarial, reconstructed and residual images in first, second and third columns respectively. First row is sampled from Tiny ImageNet and the second one is from CIFAR-10 dataset. Attack types from top to down are PGD and CW$L_2$. Note that pixel values of residual images are multiplied by a factor of $8$ for the purpose of visualization.}
    \label{fig:Reconstruction_Residual_Generation}
\end{figure}

From Table \ref{tab:REDRL} we can see compared to using only adversarial images $I_{adv}$, there is a significant boost in the accuracy of the adversarial attack recognition when the concatenation of residual and adversarial images is used. For CIFAR-10 and Tiny ImageNet datasets, the performance improvement of REDRL over attack recognition based on merely adversarial images is $63\%$ and $43\%$ respectively, showing the power of employing residuals in the estimation of sparse perturbations. Note that numbers under the ($\delta, I_{adv}$) columns show the accuracy if we had access to the ground-truth perturbations. Further, we note that the overall performance on CIFAR-10 is higher. CIFAR-10 has a much lower number of image classes and diversity. In contrast, the Tiny ImageNet has a variety of object classes and high diversity, making it more challenging for the reconstructor network $G$ to correctly model the distribution of the clean images. Another point to observe is that in the case of the Tiny ImageNet dataset, correctly classifying DeepFool, CW$L_2$, and CW$L_\infty$ attacks is more challenging for the network $\Psi$ compared to the rest of the classes. This can be attributed to the fact the victim image classifiers performance is relatively inferior compared to the ones for the CIFAR-10 dataset as reported in Table \ref{tab:image_classifier_details}. Therefore, the gradient signal back-propagating from $\mathcal{L}_{IC}$ loss, is not as informative as in the case of the CIFAR-10 dataset. Figure \ref{fig:Reconstruction_Residual_Generation} shows sample attacked images and the corresponding reconstructed and sparse residual images produced by REDRL. In addition, we would like to understand how well the reconstructor network $G$ in the REDRL pipeline can recover the original image contents from the input adversarial images. To this end, we report the performance of ResNet50, ResNext50, DenseNet121, and VGG19 image classification models on the test set samples and their respective reconstructed counterparts in Table \ref{tab:before_after}. Note that the test sets have both clean and adversarial samples. Table \ref{tab:before_after} shows the impact of the trained reconstructor $G$ by comparing classification accuracy of images before and after being applied to the reconstructor $G$. It can be observed that not only image classification accuracy significantly increases for reconstructed attacked images of different types, the accuracy on reconstructed clean samples is maintained. This is consistent across different datasets and different image classification architectures.

\subsection{Ablation Studies}
\label{subsec:ablation}

In this section, we evaluate our design choices for the proposed REDRL to better understand the contribution of Feature Reconstruction (FR) and Image Classification (IC) stages to the residual learning task. Therefore we compare the following three scenarios:
\begin{itemize}
    \item [A.] Here we ignore FR and IC stages and only optimize network $G$ for $\mathcal{L}_{R}(G)$ and $\mathcal{L}_{AC}(G)$.
    \item [B.] In this scenario, $\mathcal{L}_{F}$ is added so that network $G$ is optimized for $\mathcal{L}_{R}(G)$, $\mathcal{L}_{F}(G)$, and $\mathcal{L}_{AC}(G)$ objectives.
    \item [C.] We investigate the effect of image classification on the overall performance. Therefore, $G$ is trained for $\mathcal{L}_{R}(G)$, $\mathcal{L}_{IC}(G)$, and $\mathcal{L}_{AC}$ optimization functions.
\end{itemize}
Table \ref{tab:Ablation} presents how each step of the proposed REDRL can impact the downstream task of attack recognition. Note that in both CIFAR-10 and Tiny ImageNet datasets, REDRL consistently outperforms the rest of the scenarios in terms of overall attack detection accuracy. This shows that  enforcing FR and IC constraints simultaneously helps to learn more representative residual. It is noteworthy to mention that we observe Image Classification module in REDRL significantly contributes to the high classification score on reconstructed images. 

\begin{table}[t!]
    \centering
    \caption{Investigating the effects of Feature Reconstruction ($\mathcal{L}_{F}$) and Image Classification ($\mathcal{L}_{IC}$) objectives on the adversarial attack recognition accuracy (\%).\\}
    \resizebox{\columnwidth}{!}{%
    \begin{tabular}{c|c|c|c|c|c|c|c|c}
         \cline{2-9}
         \multicolumn{1}{c|}{} & \multicolumn{8}{|c|}{Dataset} \\
         \cline{1-9}
         \multicolumn{1}{|c|}{\multirow{2}{*}{Class}} & \multicolumn{4}{|c||}{CIFAR-10} & \multicolumn{4}{|c|}{Tiny ImageNet}\\
         \cline{2-9}
         \multicolumn{1}{|c|}{} & \multicolumn{1}{|c|}{A} & \multicolumn{1}{|c|}{B} & \multicolumn{1}{|c|}{C} & \multicolumn{1}{|c||}{REDRL} & \multicolumn{1}{|c|}{A} & \multicolumn{1}{|c|}{B} & \multicolumn{1}{|c|}{C} & \multicolumn{1}{|c|}{REDRL}
         \\
         \cline{1-9}
         \multicolumn{1}{|c|}{Clean} & \multicolumn{1}{|c|}{$99.9$} & \multicolumn{1}{|c|}{$98.9$} & \multicolumn{1}{|c|}{$100$} & \multicolumn{1}{|c||}{$100$} & \multicolumn{1}{|c|}{$99.8$} & \multicolumn{1}{|c|}{$99.5$} & \multicolumn{1}{|c|}{$99.5$} & \multicolumn{1}{|c|}{$99.7$}
         \\
         \cline{1-9}
         \multicolumn{1}{|c|}{DeepFool} & \multicolumn{1}{|c|}{$99.3$} & \multicolumn{1}{|c|}{$98.8$} & \multicolumn{1}{|c|}{$99.8$} & \multicolumn{1}{|c||}{$97.4$} & \multicolumn{1}{|c|}{$87.1$} & \multicolumn{1}{|c|}{$93.8$} & \multicolumn{1}{|c|}{$71.9$} & \multicolumn{1}{|c|}{$75.3$}
         \\
         \cline{1-9}
         \multicolumn{1}{|c|}{PGD} & \multicolumn{1}{|c|}{$99.9$} & \multicolumn{1}{|c|}{$99.6$} & \multicolumn{1}{|c|}{$99.9$} & \multicolumn{1}{|c||}{$99.9$} & \multicolumn{1}{|c|}{$99.9$} & \multicolumn{1}{|c|}{$99.8$} & \multicolumn{1}{|c|}{$99.9$} & \multicolumn{1}{|c|}{$99.9$}
         \\
         \cline{1-9}
         \multicolumn{1}{|c|}{CW$L_2$} & \multicolumn{1}{|c|}{$84.2$} & \multicolumn{1}{|c|}{$88.7$} & \multicolumn{1}{|c|}{$93.3$} & \multicolumn{1}{|c||}{$96.6$} & \multicolumn{1}{|c|}{$58.7$} & \multicolumn{1}{|c|}{$60.2$} & \multicolumn{1}{|c|}{$61.5$} & \multicolumn{1}{|c|}{$66.3$}
         \\
         \cline{1-9}
         \multicolumn{1}{|c|}{CW$L_\infty$} & \multicolumn{1}{|c|}{$63.3$} & \multicolumn{1}{|c|}{$70.8$} & \multicolumn{1}{|c|}{$71.6$} & \multicolumn{1}{|c||}{$74.1$} & \multicolumn{1}{|c|}{$42.9$} & \multicolumn{1}{|c|}{$43.0$} & \multicolumn{1}{|c|}{$53.8$} & \multicolumn{1}{|c|}{$57.7$}
         \\
         \cline{1-9}
         \multicolumn{1}{|c|}{Patch} & \multicolumn{1}{|c|}{$99.7$} & \multicolumn{1}{|c|}{$99.8$} & \multicolumn{1}{|c|}{$99.9$} & \multicolumn{1}{|c||}{$99.9$} & \multicolumn{1}{|c|}{$98.6$} & \multicolumn{1}{|c|}{$98.9$} & \multicolumn{1}{|c|}{$99.2$} & \multicolumn{1}{|c|}{$99.6$}
         \\
          \hline
         \hline
         \multicolumn{1}{|c|}{Total} & \multicolumn{1}{|c|}{$90.59$} & \multicolumn{1}{|c|}{$92.58$} & \multicolumn{1}{|c|}{$93.51$} & \multicolumn{1}{|c||}{$\textbf{94.28}$} & \multicolumn{1}{|c|}{$81.9$} & \multicolumn{1}{|c|}{$82.7$} & \multicolumn{1}{|c|}{$83.72$} & \multicolumn{1}{|c|}{$\textbf{85.57}$}
         \\
         \cline{1-9}
    \end{tabular}
    }
    \label{tab:Ablation}
\end{table}
\section{Saliency Profiles}
\label{sec:saliency}
\begin{figure*}[t!]
    \centering
    \subfloat[][Clean]{\includegraphics[width=0.5\textwidth]{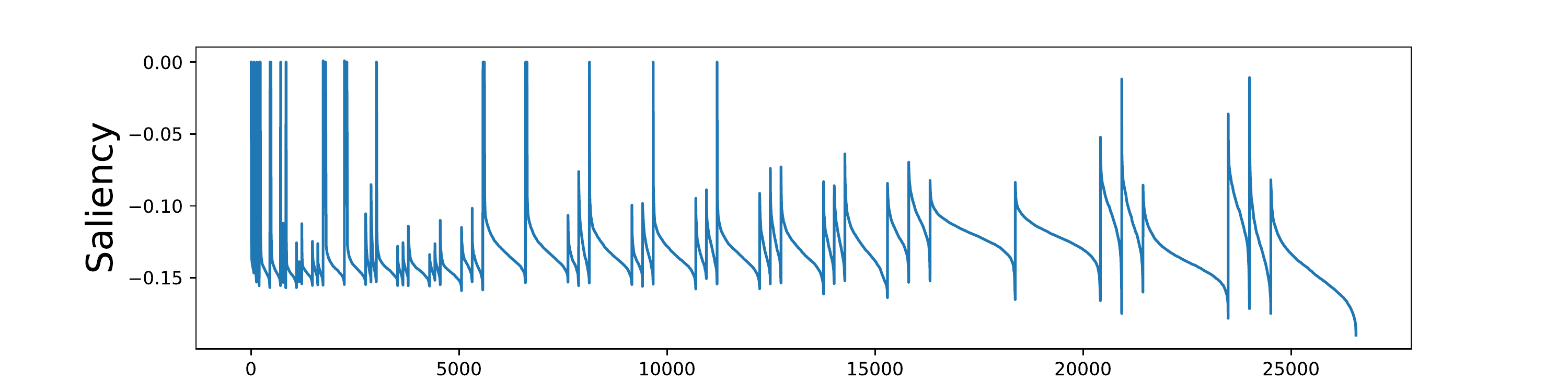}}~
    \subfloat[][CW$L_2$]{\includegraphics[width=0.5\textwidth]{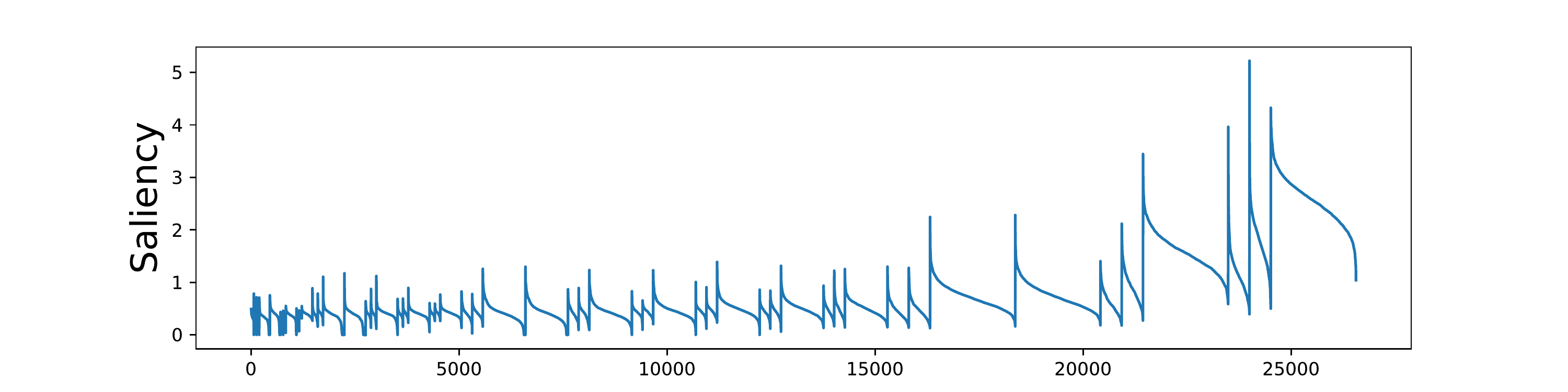}}\\\vspace{-.3cm}
    \subfloat[][CW$L_\infty$]{\includegraphics[width=0.5\textwidth]{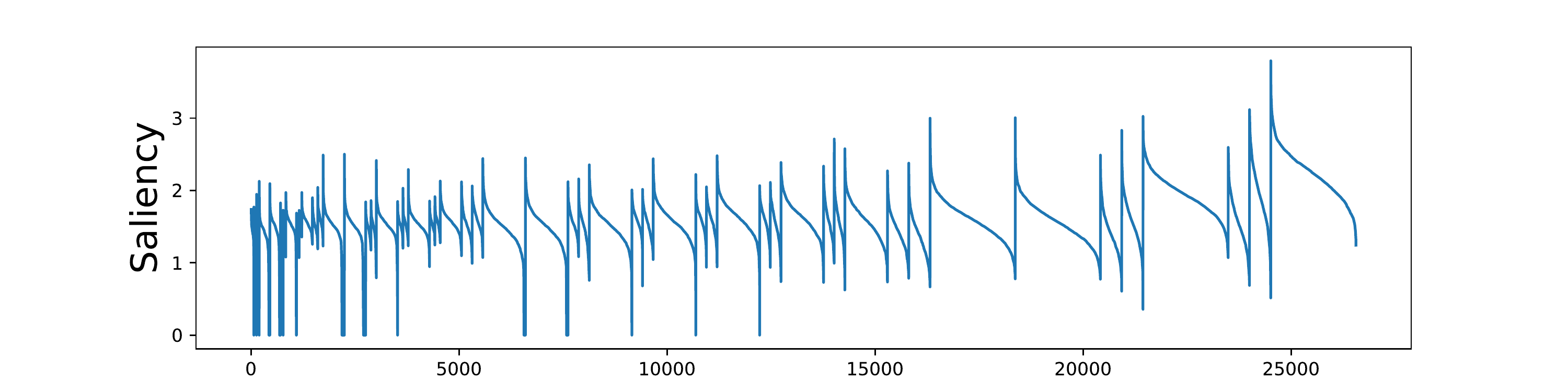}}~
    \subfloat[][PGD]{\includegraphics[width=0.5\textwidth]{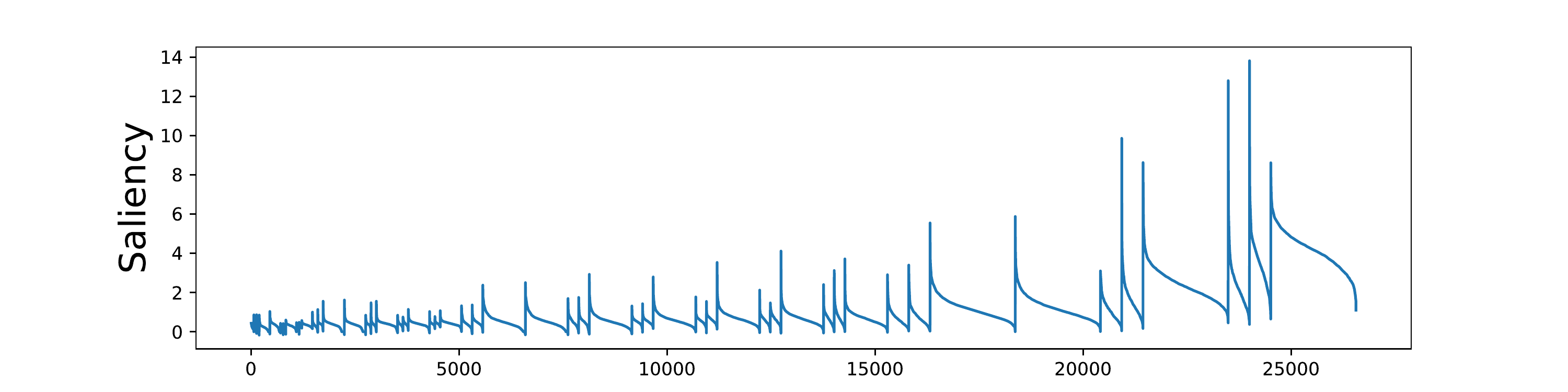}}\\\vspace{-.3cm}
    \subfloat[][Patch]{\includegraphics[width=0.5\textwidth]{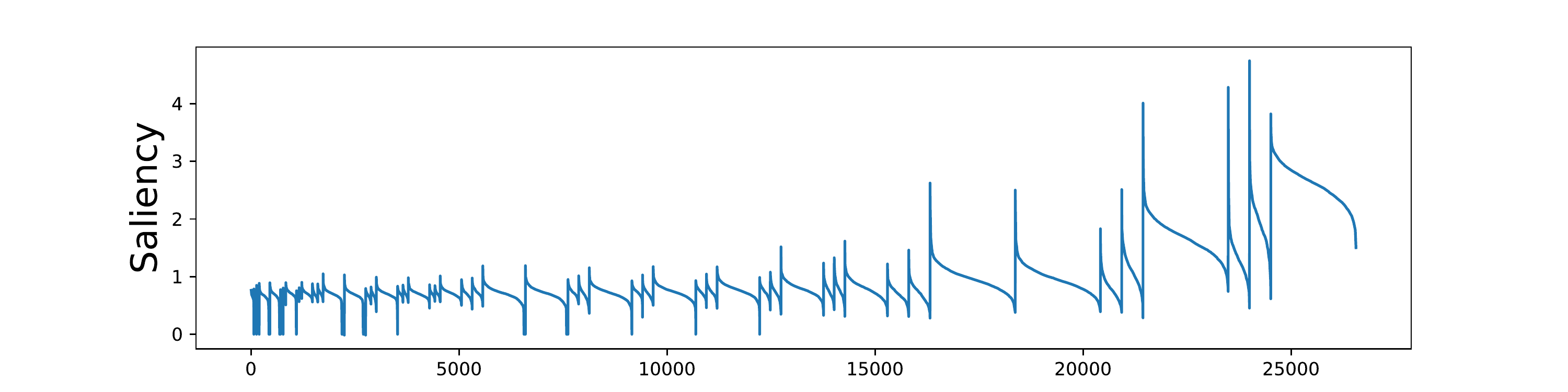}}~
    \subfloat[][DeepFool]{\includegraphics[width=0.5\textwidth]{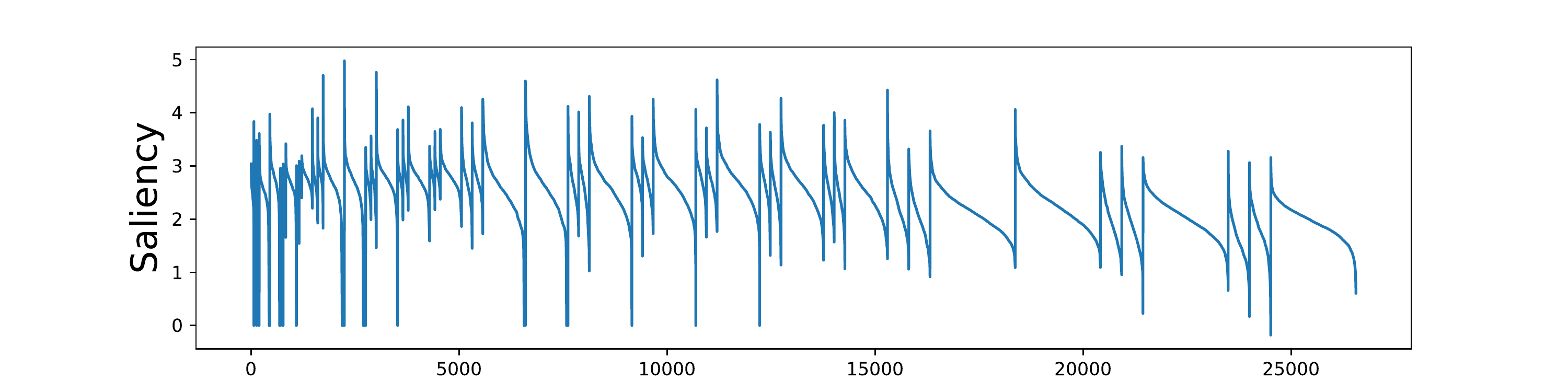}}\\\vspace{0.5cm}
    \caption{Parameter saliency profiles for different adversarial attacks on CIFAR-10 and using ResNet50 image classifier. Perturbations used for PGD and CW$L_\infty$ have $L_\infty$-norm bounded by $16/255$. Note the magnitude of saliency profiles on y-axis when comparing different attacks. Details of different attacks are presented in supplementary material.}
    \label{fig:saliency_profiles_main_1}
\end{figure*}
\begin{figure*}[t!]
    \centering
    \subfloat[][Original Image]{\includegraphics[width=0.12\textwidth]{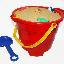}}~
    \subfloat[][CW$L_2$]{\includegraphics[width=0.12\textwidth]{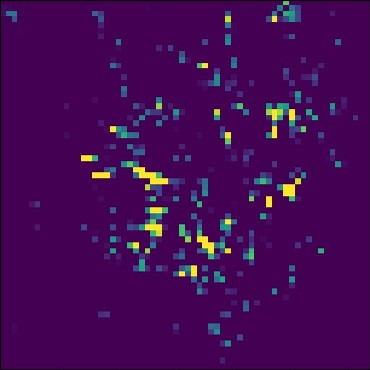}}~
    \subfloat[][CW$L_\infty$]{\includegraphics[width=0.12\textwidth]{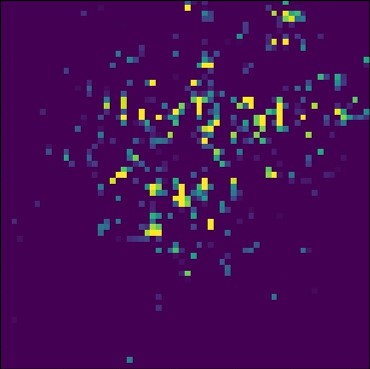}}~
    \subfloat[][DeepFool]{\includegraphics[width=0.12\textwidth]{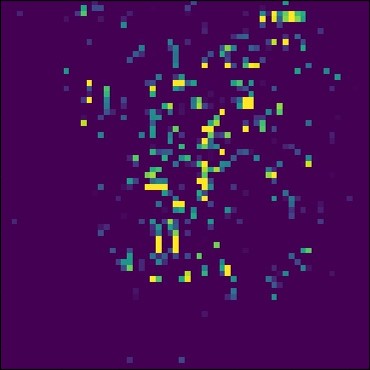}}~
    \subfloat[][Patch]{\includegraphics[width=0.12\textwidth]{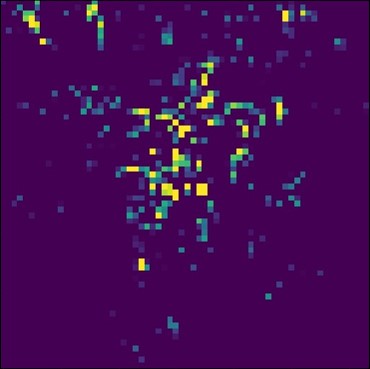}}~
    \subfloat[][PGD]{\includegraphics[width=0.12\textwidth]{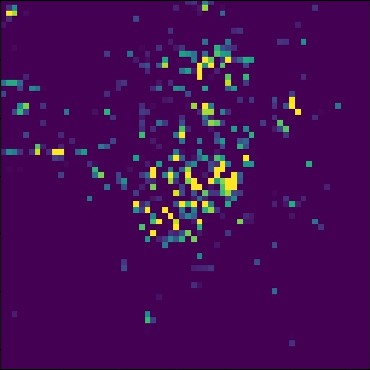}}\\\vspace{0.5cm}
    
    \caption{Input saliency heat-maps for different adversarial attacks on Tiny ImageNet using ResNet50 image classifier. Perturbations used for PGD and CW$L_\infty$ have $L_\infty$-norm bounded by $16/255$. We can observe from the figures that each attack has a specific saliency heat-map. Details of different attacks are presented in supplementary material.}
    \label{fig:heat_maps_main_1}
\end{figure*}
In the previous section, we observed that adversarial attack algorithms generate distinct and identifiable patterns.  This observation raises the question, do different attacks yield qualitatively different downstream behavior in neural networks? To answer this question, we make use of recent work in parameter-space saliency maps \cite{levin2021models}.
 
\subsection{Parameter-Space Saliency Maps}
One intuitive approach to parameter space saliency would involve simply obtaining the gradient of the loss with respect to the parameters on a sample of interest and computing its entry-wise absolute value.  The authors of \cite{levin2021models} note that early layers typically exhibit higher saliency under this definition since their filters extract broadly applicable low-level features and errors propagated from early layers accumulate during a forward pass, yet this phenomenon does not imply that early filters are responsible for misbehavior specific to the sample of interest.  Thus, they standardize the saliency values across the entire validation set so that saliency for a particular parameter now represents the number of standard deviations away from the mean saliency value (of that parameter) over validation samples.  Additionally, the authors note that individual filters are often responsible for interpretable tasks, so they aggregate saliency by filter.  Finally, saliency can be written as
\begin{equation}
    \hat{s}_{k}(x, y) := \frac{|\frac{1}{|\alpha_k|}\sum_{i \in \alpha_k} |\nabla_{\theta_{i}} \mathcal{L}_{\theta}(x, y)|-\mu_{k}|}{\sigma_{k}},
\end{equation}
where $\mu_k$ and $\sigma_k$ denote the mean and standard deviation, respectively, of $\frac{1}{|\alpha_k|}\sum_{i \in \alpha_k} |\nabla_{\theta_{i}} \mathcal{L}_{\theta}(x, y)|$ across all $(x, y)$ pairs in the validation set, and $\alpha_k$ denotes all indices of parameters belonging to filter $k$.  We use this definition of saliency in our subsequent experiments.  By plotting such saliency maps, we can identify exactly where in the network an adversarial attack triggers malicious behavior.
 
The authors of  \cite{levin2021models} further use this method to compute input-space saliency maps, $M_F = |\nabla_x D_{C}(\hat{s}(x,y), \hat{s}')|$, where $D_{C}$ denotes cosine distance, and $\hat{s}'$ is a copy of $\hat{s}(x,y)$ in which entries corresponding to the top $n$ most salient filters are boosted.  We will also use these input-space saliency maps to analyze the distinct patterns of adversarial attacks.  These input-space maps indicate exactly which input features are triggering the misbehaviors identified in parameter space.

Figure \ref{fig:saliency_profiles_main_1} contains the saliency profiles of clean images and adversarial examples on the CIFAR-10 dataset using a ResNet50 model, where filters in each layer are sorted from most to least salient as in \cite{levin2021models} and then the profiles are averaged over the entire test set. We observe that each attack has a unique and dramatically different saliency map. For example, we see that PGD, CW$L_2$, and the Patch attack mostly target the final layers, while DeepFool focuses on the earlier layers, and CW$L_\infty$ targets almost all layers. These observations are consistent across different datasets and architectures (see supplementary materials for additional saliency profiles). Note that even though PGD, CW$L_2$, and Adversarial Patch seem to yield similar saliency profiles, they have different magnitudes, \emph{i.e.} different values on the y-axis.


Figure \ref{fig:heat_maps_main_1} depicts the previously described input-space visualizations derived from the parameter-space saliency maps.  We see that different attacks highlight distinct image-space regions.  This observation indicates that the model's filters which are triggered most by an attack, perform different functions across the various attacks we examine.  We conclude from these heat-maps that different attacks have qualitatively different impacts on the behavior of the victims models.
\section{Conclusion}
\label{conclusion}


Adversarial attack algorithms are historically compared only by their success rates at fooling victim models.  We find that these attacks produce distinct perturbation patterns which can be classified from the adversarial example alone (without access to the perturbation itself).  In order to do so, we present a novel end-to-end framework, REDRL, for extracting adversarial perturbations from perturbed images and classifying the parent adversary. Moreover, we use parameter-space saliency profiles to investigate the downstream effects of adversarial attacks on their victims and find unique attack-specific behavior. Namely, some attacks (\emph{e.g.}, PGD) trigger misbehavior in deeper components of the victim architecture, while others (\emph{e.g.}, DeepFool) primarily target shallow components.  We conclude from these exploratory experiments that adversarial examples do not simply differ in effectiveness, and our work leaves open the possibility that practitioners deciding on an attack for their use-case should consider qualitative impacts other than simply the success rate at degrading test accuracy.

{\small
\bibliographystyle{ieee_fullname}
\bibliography{main}
}

\newpage
\section*{Appendix}
\renewcommand{\thesection}{\Alph{section}}

\setcounter{section}{0}

\section{Adversarial Attack Configurations}
As mentioned in the paper, we adopted five adversarial attack models, namely, DeepFool, CW$L_2$, CW$L_\infty$, PGD and Adversarial Patch. Therefore, we created five different adversarial versions of CIFAR-10 and Tiny ImageNet datasets for the purpose of our experiments. Table \ref{tab:attack_configs_supp} summarizes the configurations and hyper-parameters we used for each attack to create the respective adversarial datasets. 

\begin{table}[!htb]
    \centering
    \caption{Adversarial Attacks Settings}
     \resizebox{\columnwidth}{!}{%
    \begin{tabular}{|c|c|}
    \hline
    \multicolumn{1}{|c|}{Attack Type} & \multicolumn{1}{|c|}{Configuration} \\
    \hline
    \hline
    \multicolumn{1}{|c|}{DeepFool} & Steps: $50$ \\
    \hline
    \hline
    \multicolumn{1}{|c}{\multirow{2}{*}{PGD}} & \multicolumn{1}{|c|}{$\epsilon \in \{4,8,16\}$} \\
    \cline{2-2}
    \multicolumn{1}{|c}{} & \multicolumn{1}{|c|}{$\alpha: 0.01$, Steps: $100$} \\
    \hline
    \hline
    \multicolumn{1}{|c}{\multirow{2}{*}{CW$L_2$}} &  \multicolumn{1}{|c|}{Steps: $1000$, $c \in \{100, 1000\}$} \\
    \cline{2-2}
    \multicolumn{1}{|c}{} & \multicolumn{1}{|c|}{Learning Rate: $0.01$, $\kappa : 0$} \\
    \hline
    \hline
    \multicolumn{1}{|c}{\multirow{2}{*}{CW$L_\infty$}} &  \multicolumn{1}{|c|}{Steps: $100$, $\epsilon \in \{4,8,16\}$} \\
    \cline{2-2}
    \multicolumn{1}{|c}{} & \multicolumn{1}{|c|}{Learning Rate: $0.005$, c: $5$} \\
    \hline
    \hline
    \multicolumn{1}{|c}{\multirow{2}{*}{Adversarial Patch}} & \multicolumn{1}{|c|}{Steps: $100$, $\epsilon \in \{4,8,16\}$} \\
    \cline{2-2}
    \multicolumn{1}{|c}{} & \multicolumn{1}{|c|}{Patch Size $\in \{4\times4, 8\times8, 16 \times 16\}$} \\
    \hline
    \end{tabular}
    }
    \label{tab:attack_configs_supp}
\end{table}

In Table \ref{tab:attack_configs_supp}, Steps represent the number of iterations for which an adversarial perturbation was optimized to maximize the loss an image classification model. In addition, $\epsilon$ and $\alpha$ are the maximum allowable perturbation size and the step size in each iteration respectively. For CW$L_2$, $c$ is the hyper-parameter which shows the trade-off between the distance and  confidence terms in the optimization objective; also $\kappa$ represents the confidence of adversarial examples.

\section{Saliency Profile}
\subsection{Input-Space Saliency Map}
Due to the space constraint, we did not incorporate an example of input-saliency map for CIFAR10 dataset. However, here we show an example of input-pace saliency maps for an sample of CIFAR-10 dataset using ResNext50 model.

\begin{figure}[t!]
    \centering
    \subfloat[][Original Image]{\includegraphics[width=0.12\textwidth]{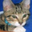}}~
    \subfloat[][CW$L_2$]{\includegraphics[width=0.12\textwidth]{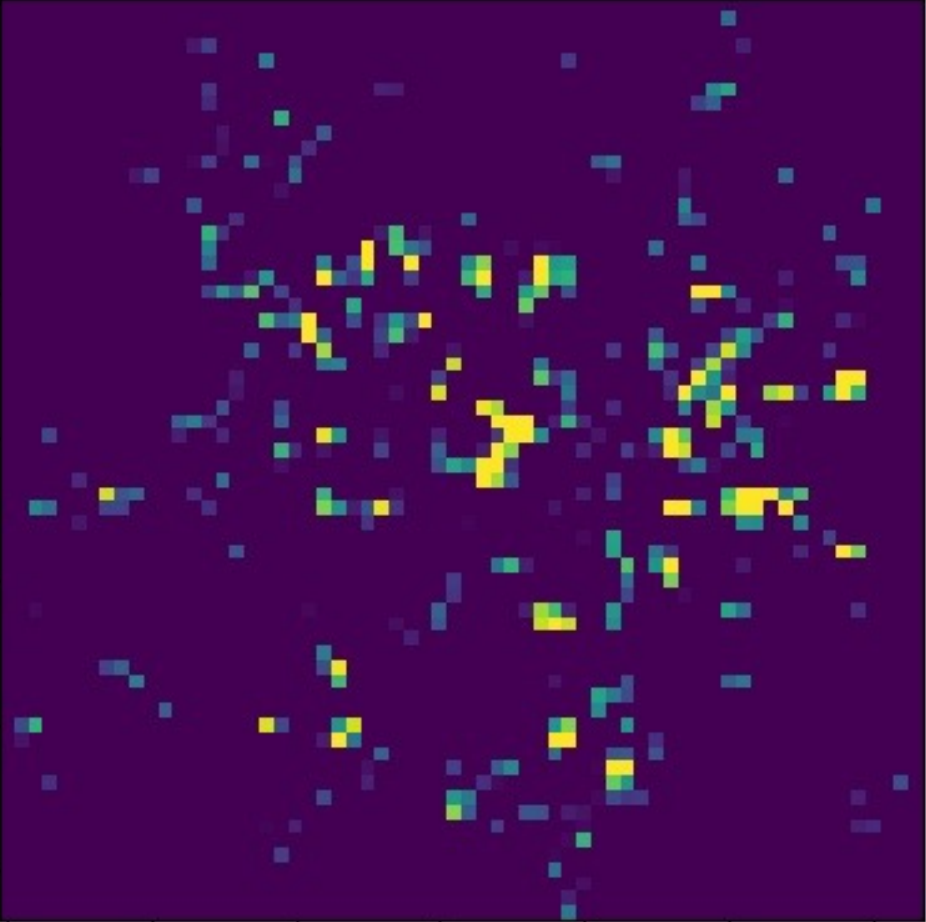}}~
    \subfloat[][CW$L_\infty$]{\includegraphics[width=0.12\textwidth]{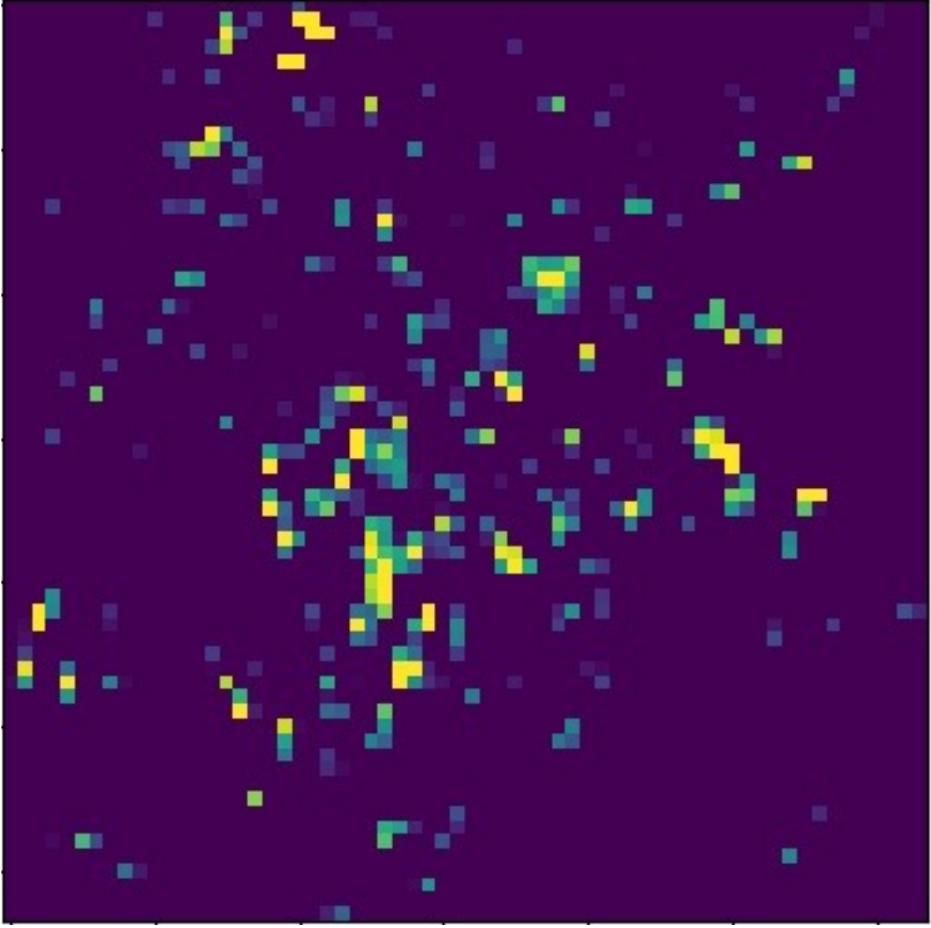}}\\
    \subfloat[][DeepFool]{\includegraphics[width=0.12\textwidth]{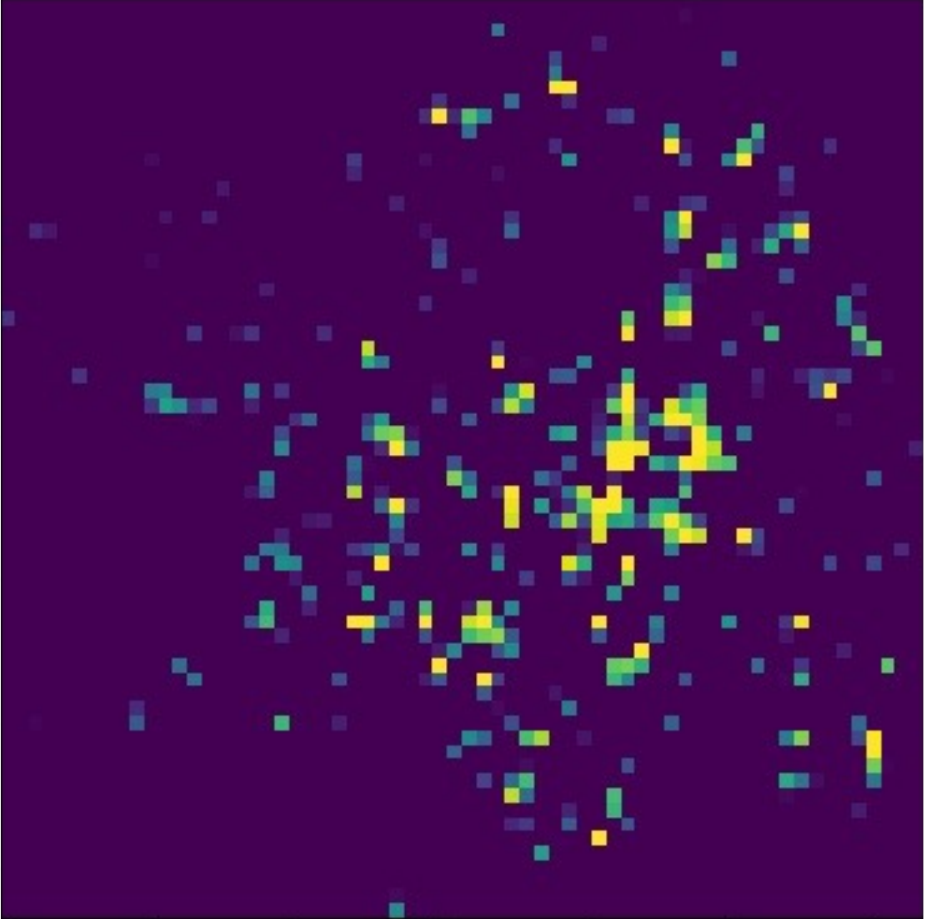}}~
    \subfloat[][Patch]{\includegraphics[width=0.12\textwidth]{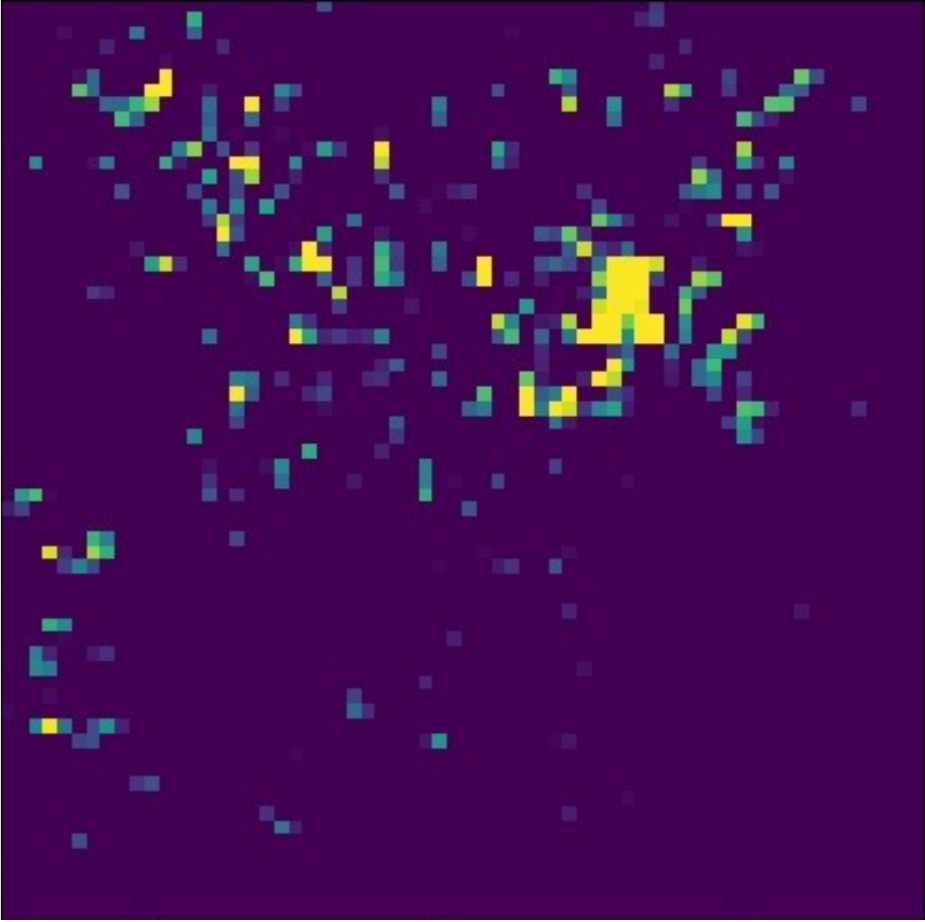}}~
    \subfloat[][PGD]{\includegraphics[width=0.12\textwidth]{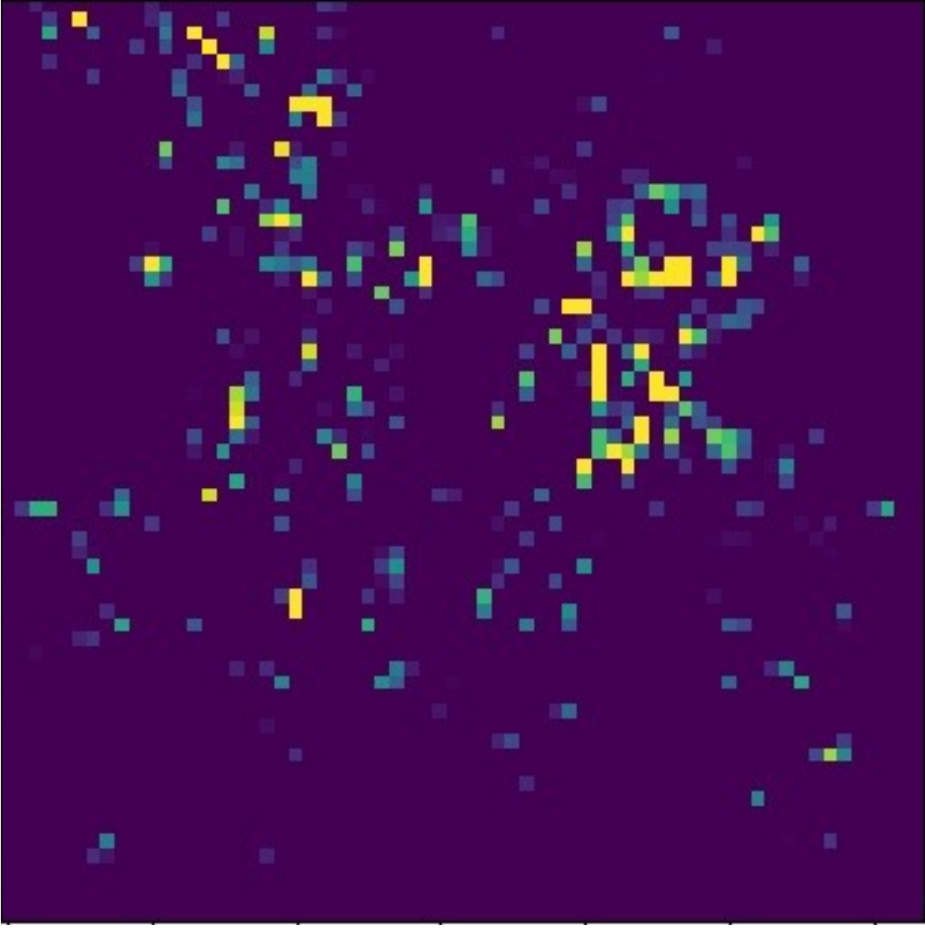}}\\\vspace{0.5cm}
    
    \caption{Input saliency heat-maps for different adversarial attacks on CIFAR-10 dataset using ResNet50 image classifier. We can observe from the figures that each attack has a specific saliency heat-map}
    \label{fig:heat_maps_cifar_resnext50}
\end{figure}

Similar to Figure 6 in the paper, from Figure \ref{fig:heat_maps_cifar_resnext50} we can see that different attack algorithms have different input-saliency maps despite using a fixed image classification model.

\begin{figure*}[t!]
    \centering
    \subfloat[][Clean]{\includegraphics[width=0.5\textwidth]{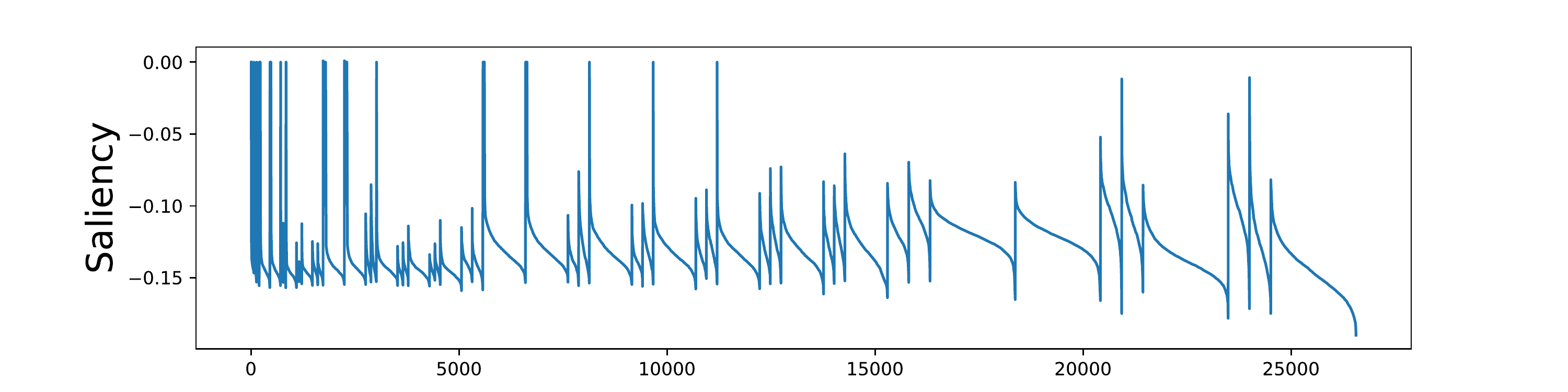}}~
    \subfloat[][CW$L_2$]{\includegraphics[width=0.5\textwidth]{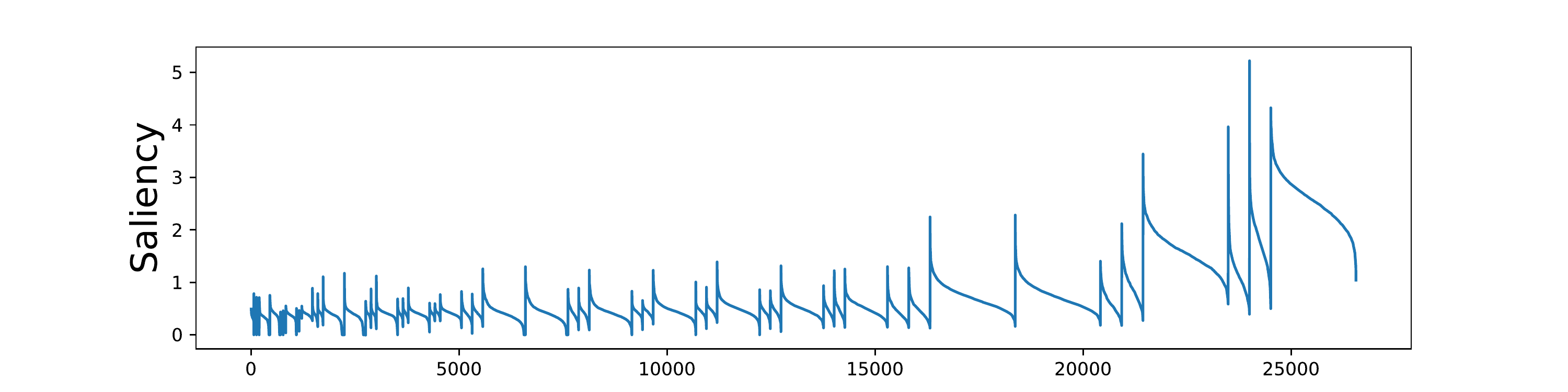}}\\\vspace{-.3cm}
    \subfloat[][CW$L_\infty$]{\includegraphics[width=0.5\textwidth]{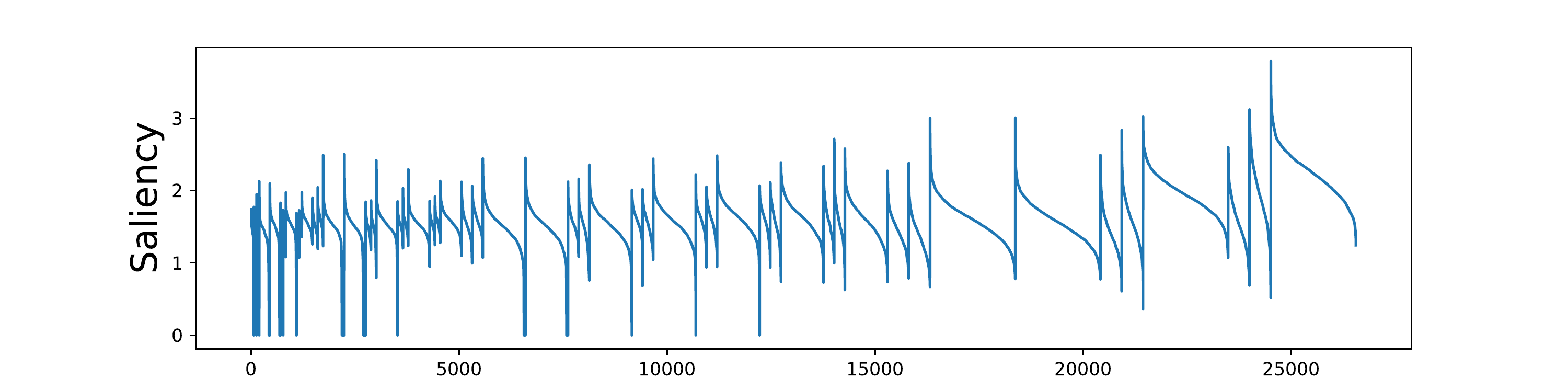}}~
    \subfloat[][PGD]{\includegraphics[width=0.5\textwidth]{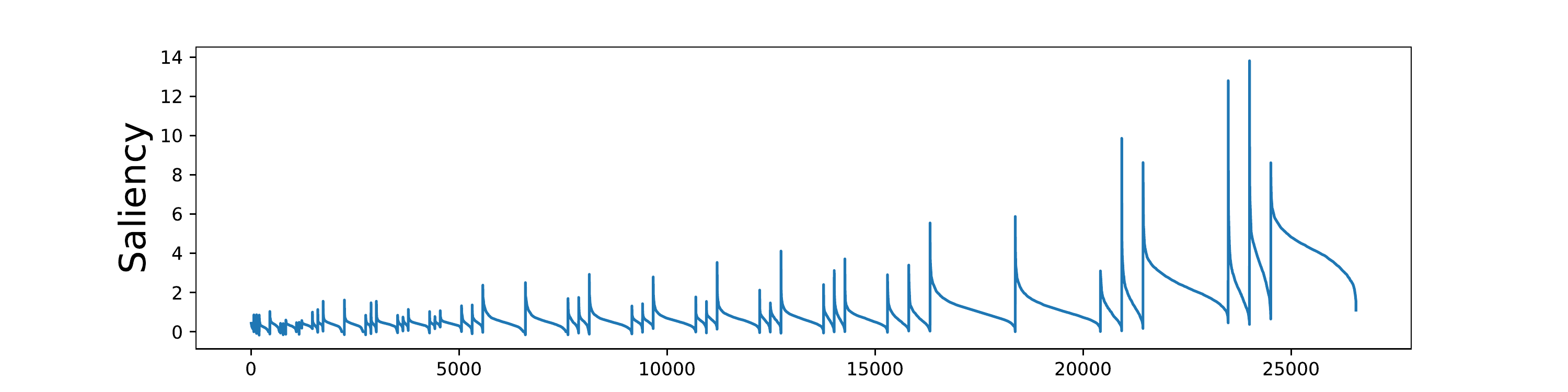}}\\\vspace{-.3cm}
    \subfloat[][Patch]{\includegraphics[width=0.5\textwidth]{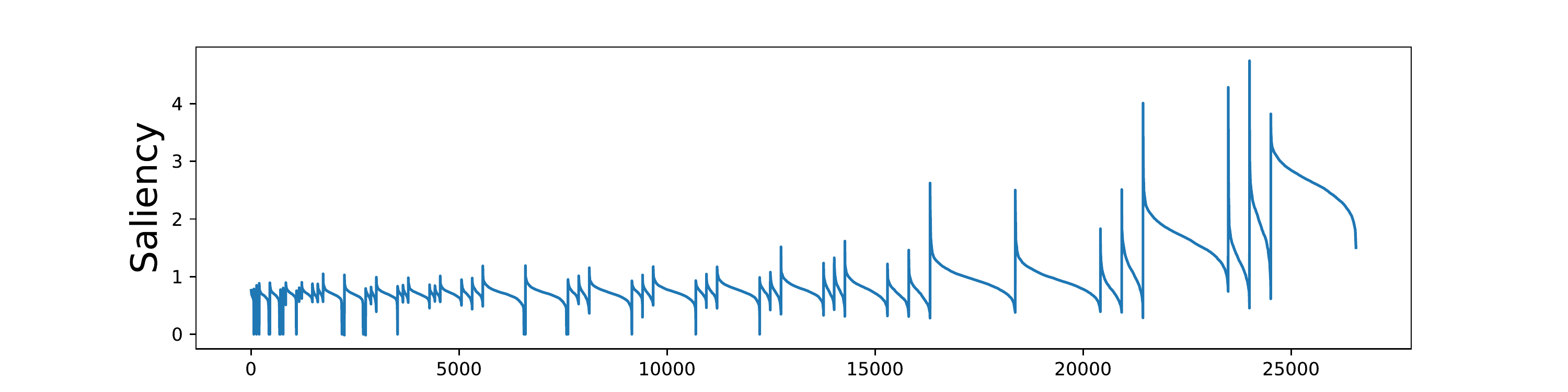}}~
    \subfloat[][DeepFool]{\includegraphics[width=0.5\textwidth]{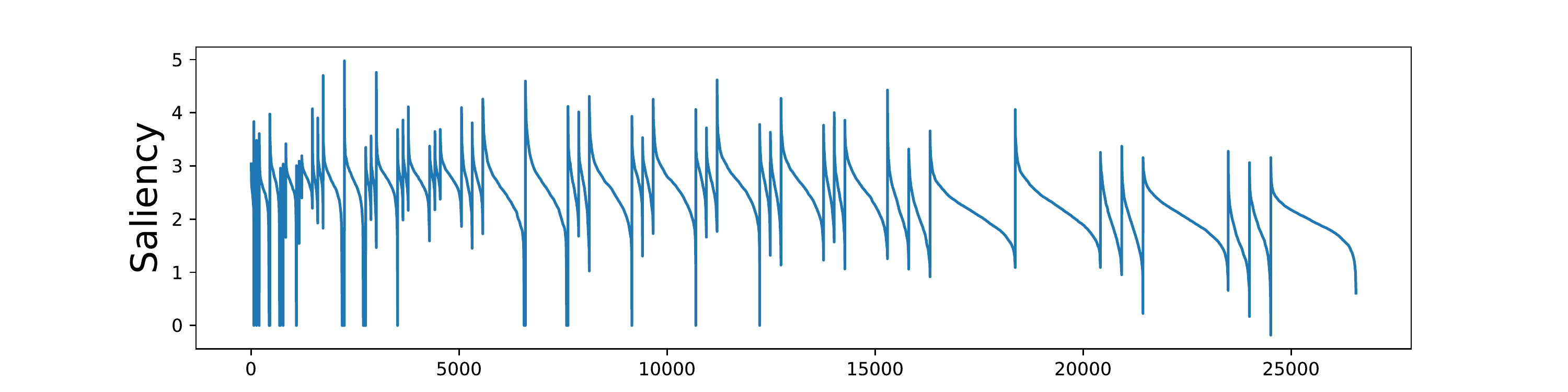}}\\\vspace{0.5cm}
    \caption{Parameter-space saliency profiles for different adversarial attacks on CIFAR-10 and using ResNet50 image classifier. Note the magnitude of saliency profiles on y-axis when comparing different attacks.}
    \label{fig:saliency_profiles_cifar10_resnet50}
\end{figure*}
\begin{figure*}[t!]
    \centering
    \subfloat[][Clean]{\includegraphics[width=0.5\textwidth]{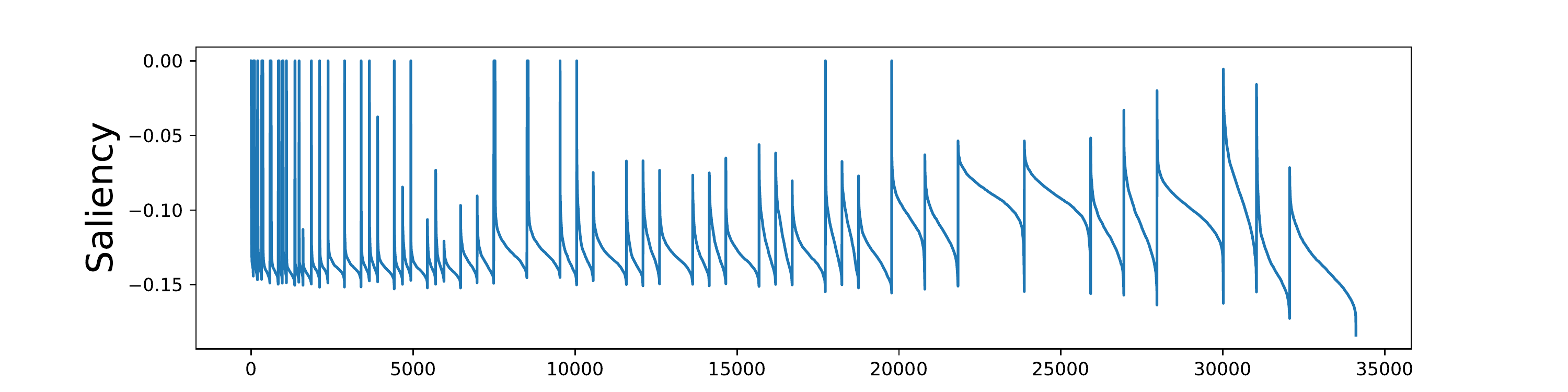}}~
    \subfloat[][CW$L_2$]{\includegraphics[width=0.5\textwidth]{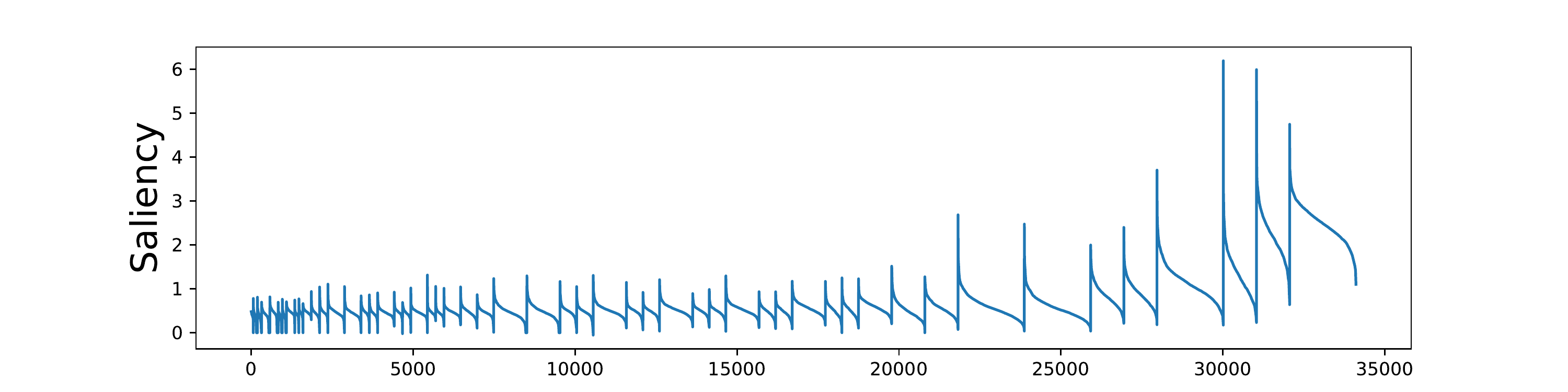}}\\\vspace{-.3cm}
    \subfloat[][CW$L_\infty$]{\includegraphics[width=0.5\textwidth]{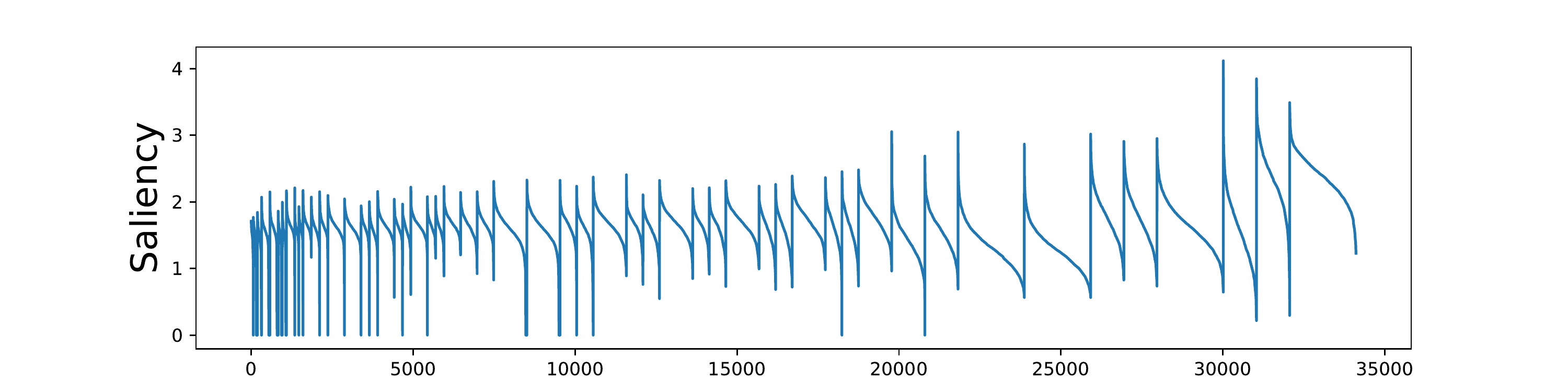}}~
    \subfloat[][PGD]{\includegraphics[width=0.5\textwidth]{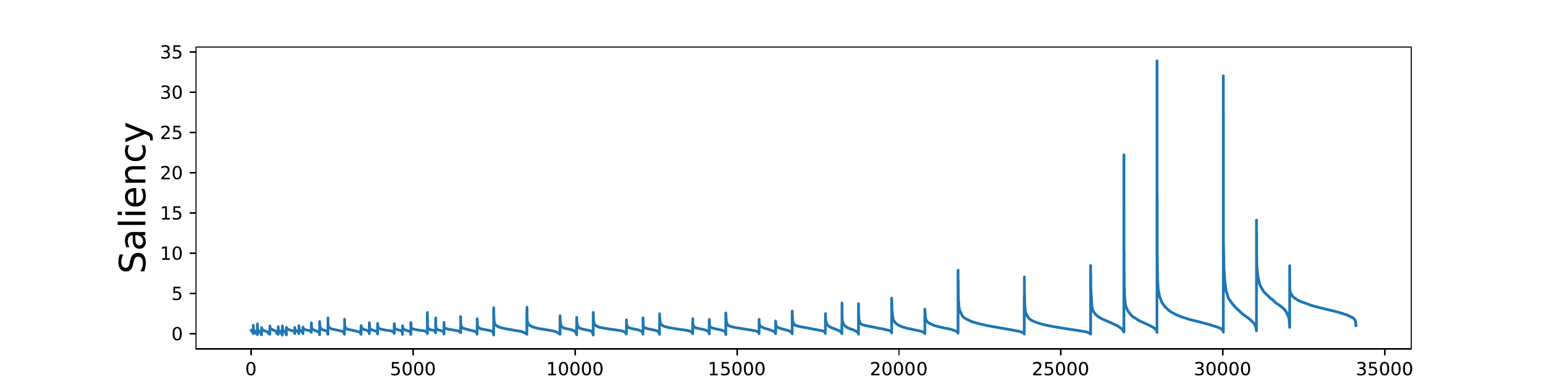}}\\\vspace{-.3cm}
    \subfloat[][Patch]{\includegraphics[width=0.5\textwidth]{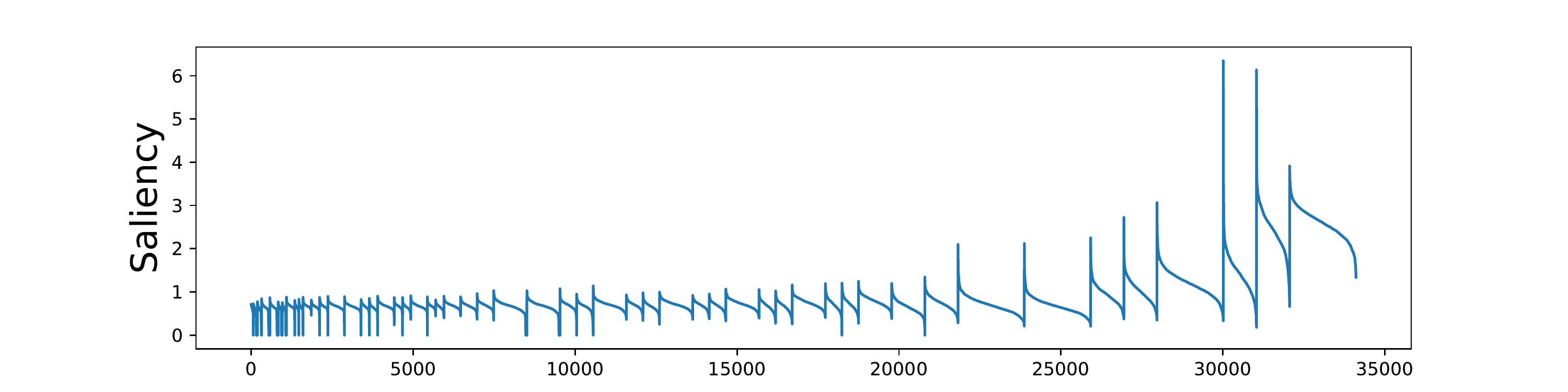}}~
    \subfloat[][DeepFool]{\includegraphics[width=0.5\textwidth]{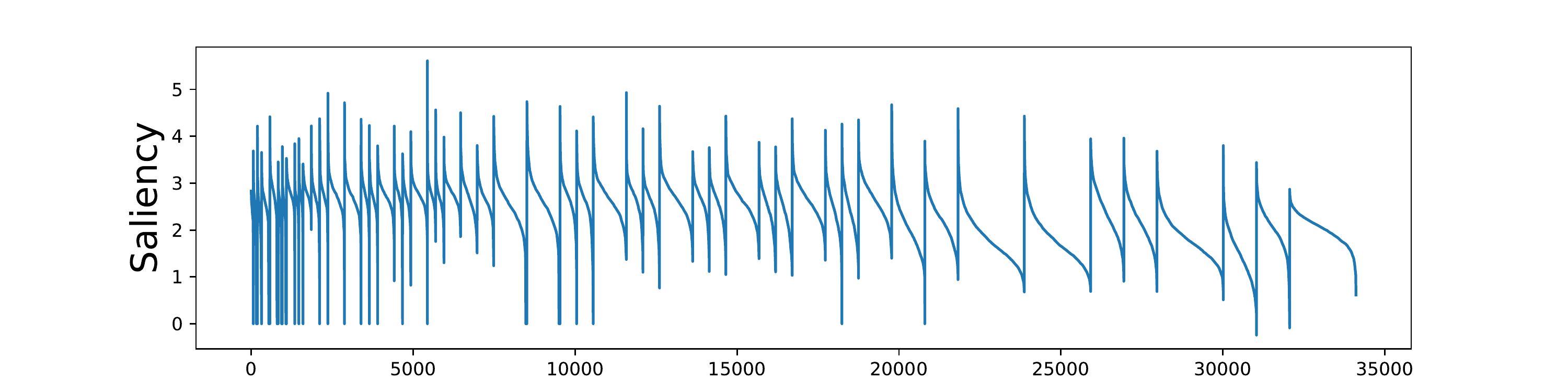}}\\\vspace{0.5cm}
    \caption{Parameter-space saliency profiles for different adversarial attacks on CIFAR-10 and using ResNext50 image classifier. Note the magnitude of saliency profiles on y-axis when comparing different attacks.}
    \label{fig:saliency_profiles_cifar10_resnext50}
\end{figure*}
\begin{figure*}[t!]
    \centering
    \subfloat[][Clean]{\includegraphics[width=0.5\textwidth]{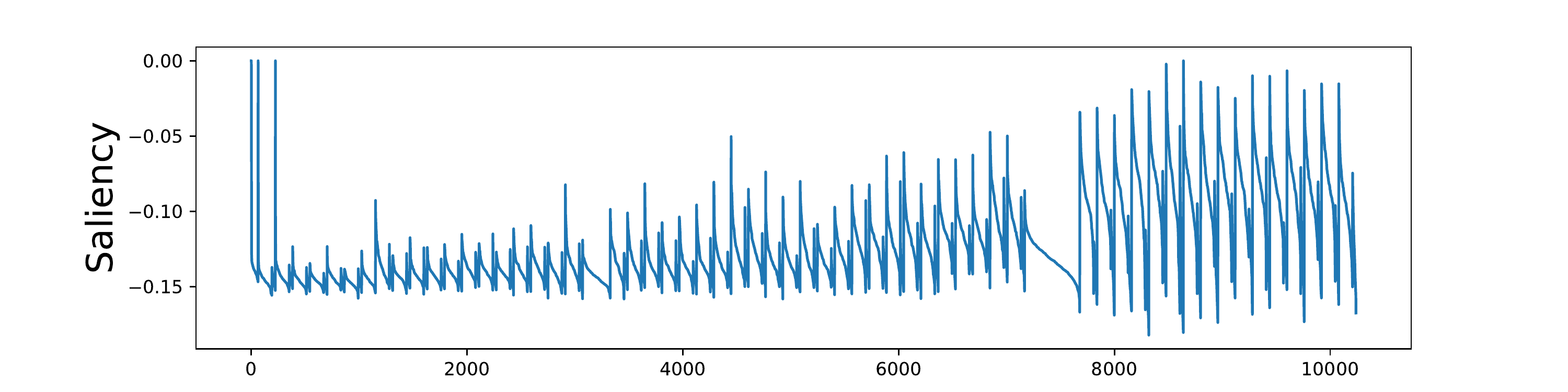}}~
    \subfloat[][CW$L_2$]{\includegraphics[width=0.5\textwidth]{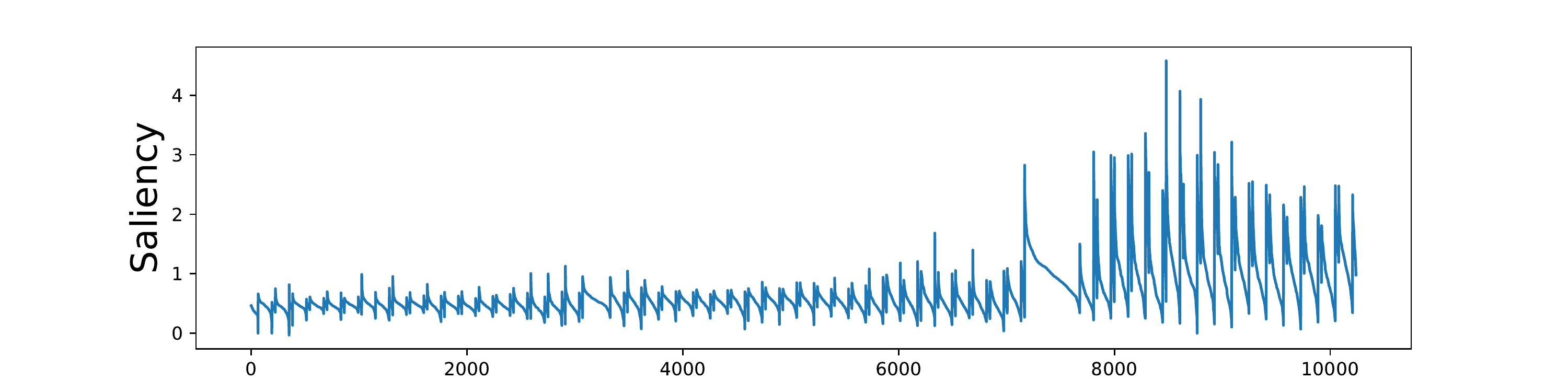}}\\\vspace{-.3cm}
    \subfloat[][CW$L_\infty$]{\includegraphics[width=0.5\textwidth]{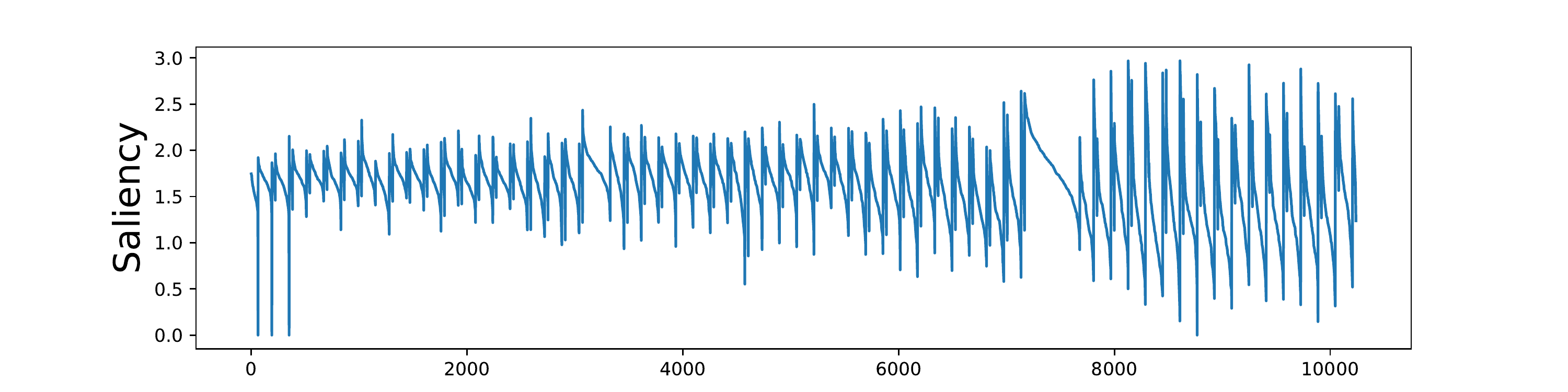}}~
    \subfloat[][PGD]{\includegraphics[width=0.5\textwidth]{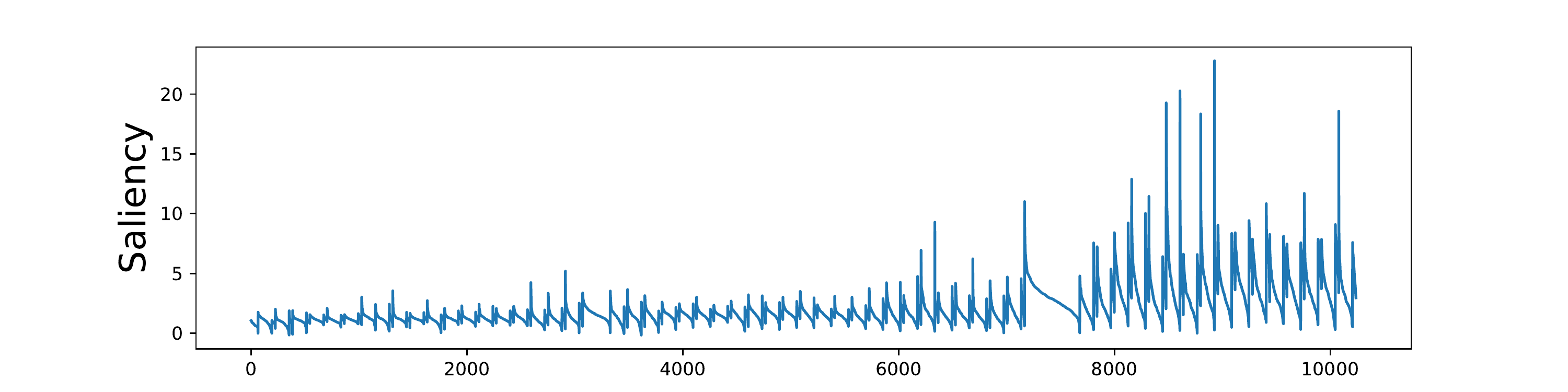}}\\\vspace{-.3cm}
    \subfloat[][Patch]{\includegraphics[width=0.5\textwidth]{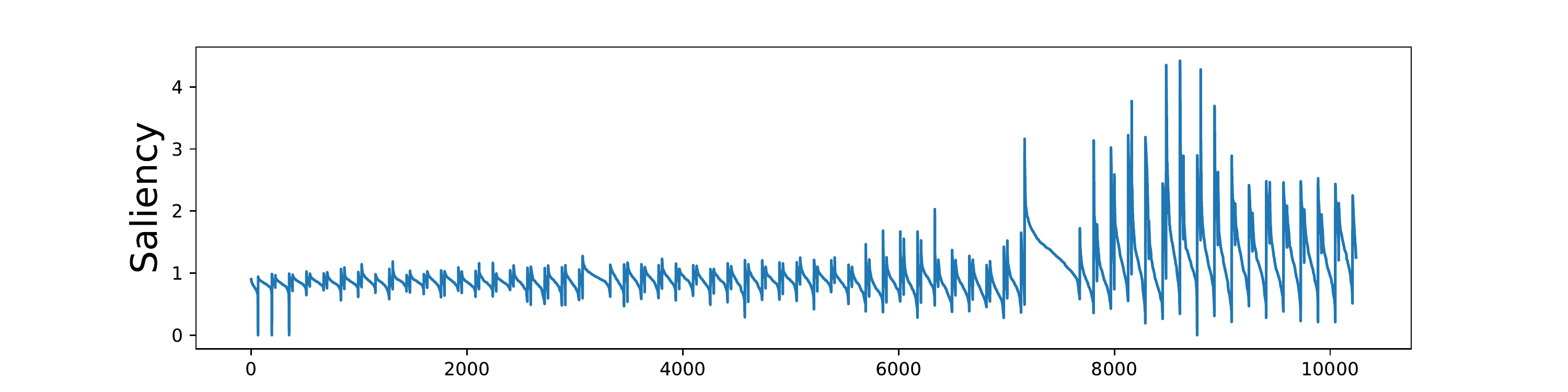}}~
    \subfloat[][DeepFool]{\includegraphics[width=0.5\textwidth]{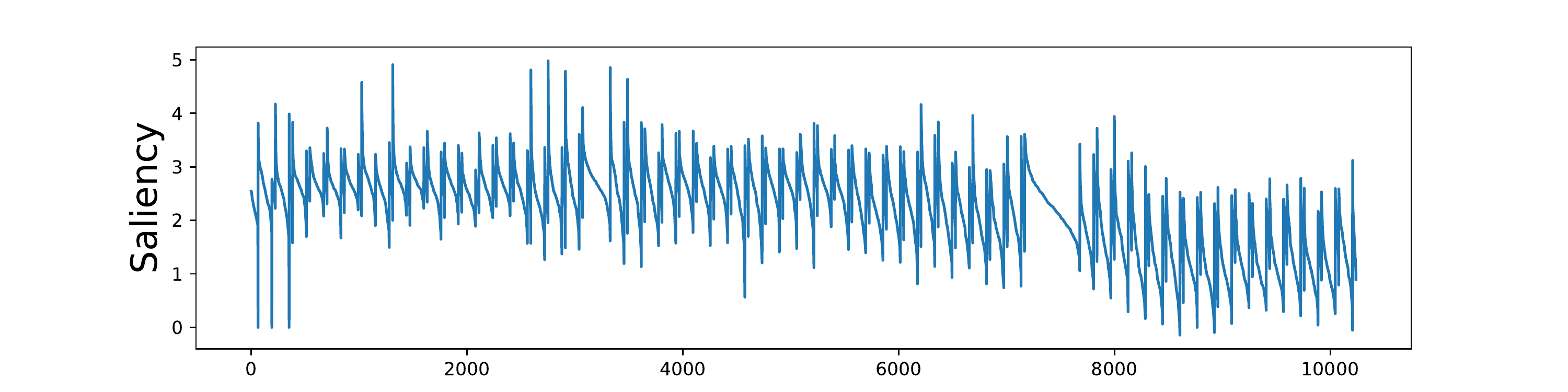}}\\\vspace{0.5cm}
    \caption{Parameter-space saliency profiles for different adversarial attacks on CIFAR-10 and using DenseNet121 image classifier. Note the magnitude of saliency profiles on y-axis when comparing different attacks.}
    \label{fig:saliency_profiles_cifar10_denseset121}
\end{figure*}
\begin{figure*}[t!]
    \centering
    \subfloat[][Clean]{\includegraphics[width=0.5\textwidth]{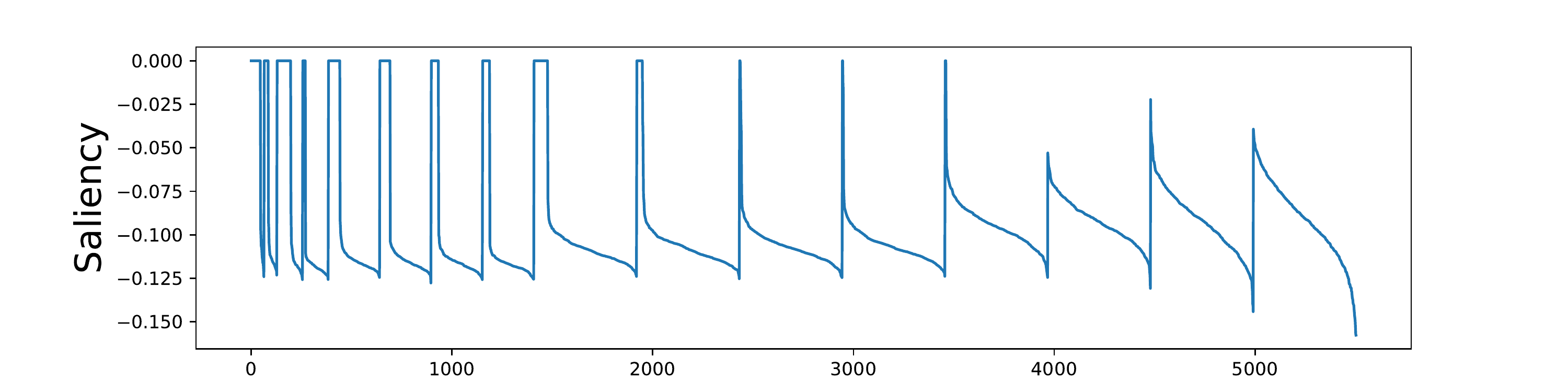}}~
    \subfloat[][CW$L_2$]{\includegraphics[width=0.5\textwidth]{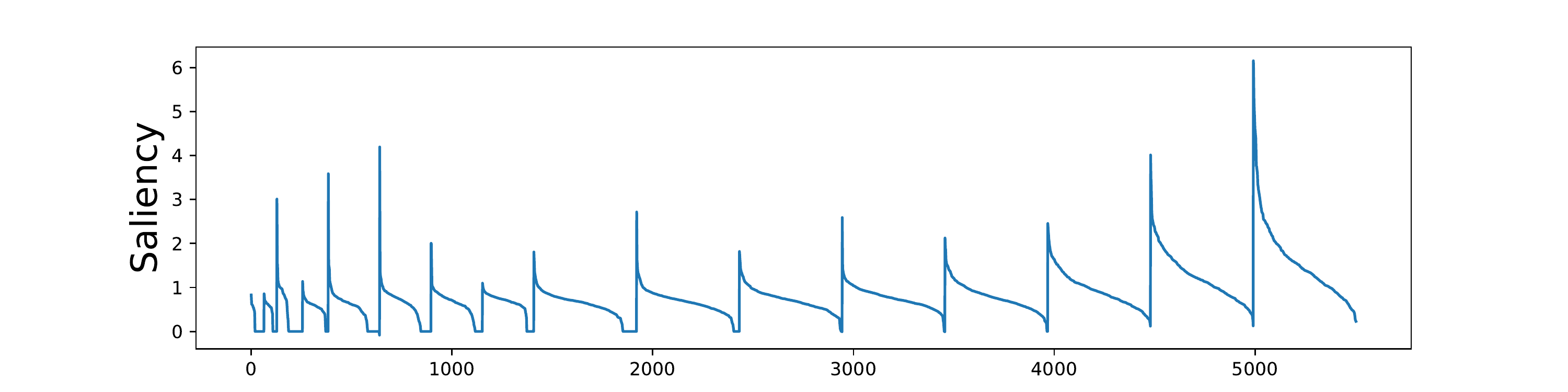}}\\\vspace{-.3cm}
    \subfloat[][CW$L_\infty$]{\includegraphics[width=0.5\textwidth]{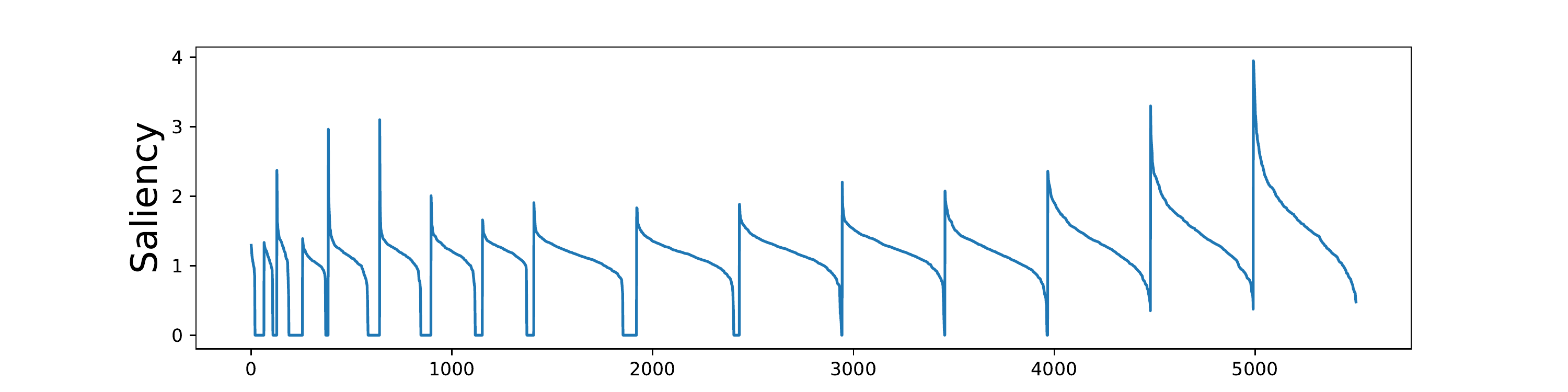}}~
    \subfloat[][PGD]{\includegraphics[width=0.5\textwidth]{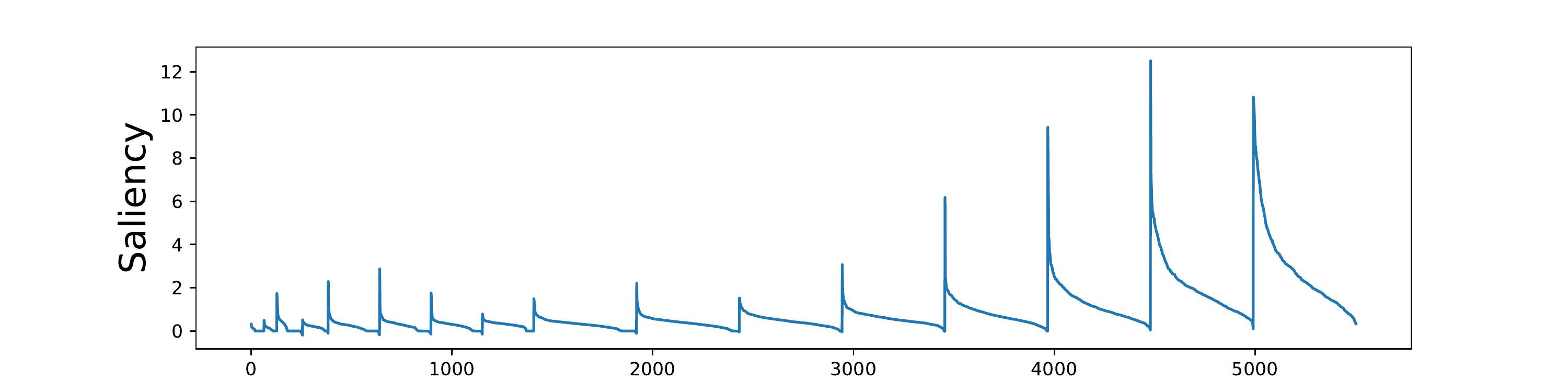}}\\\vspace{-.3cm}
    \subfloat[][Patch]{\includegraphics[width=0.5\textwidth]{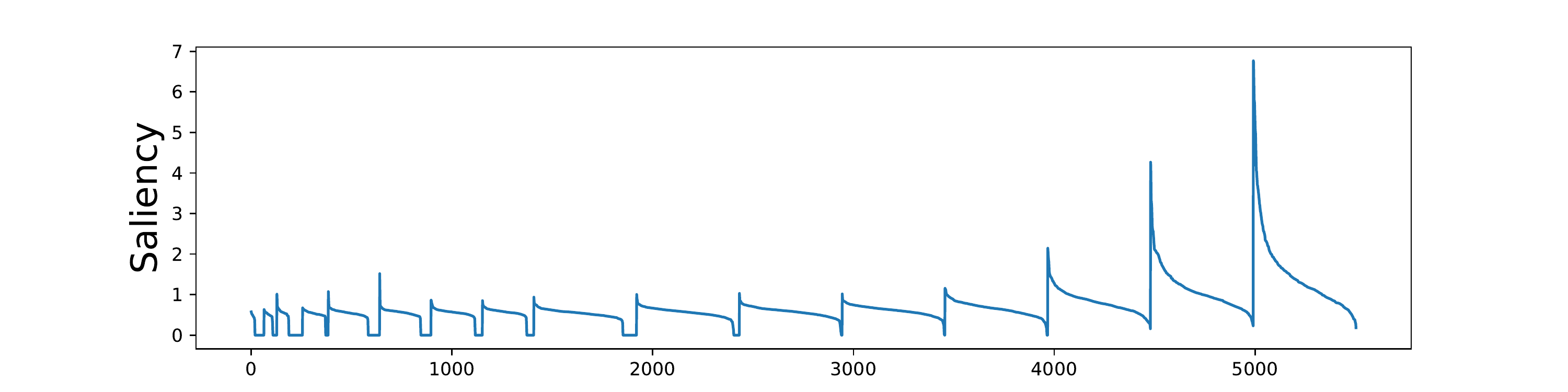}}~
    \subfloat[][DeepFool]{\includegraphics[width=0.5\textwidth]{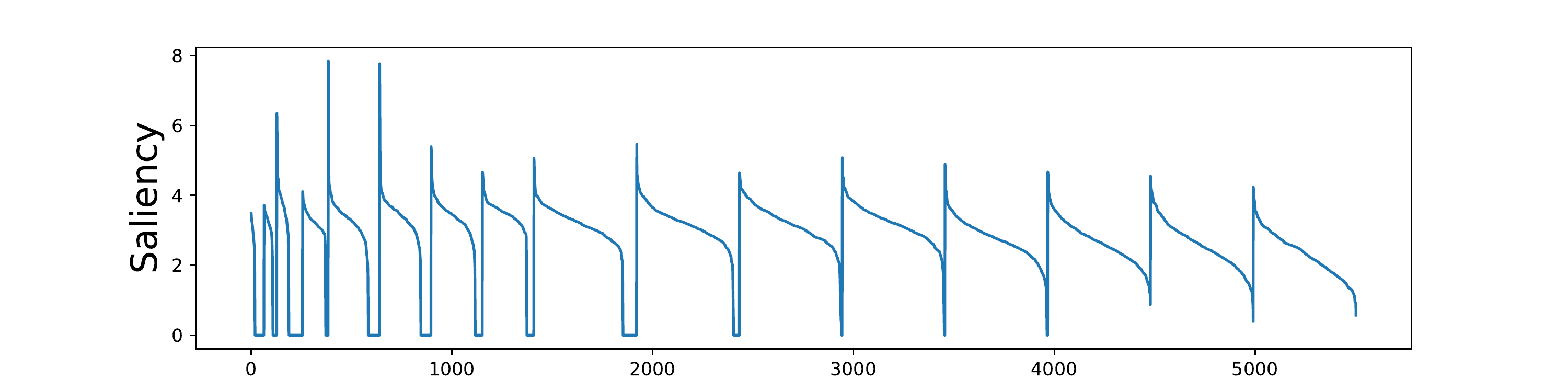}}\\\vspace{0.5cm}
    \caption{Parameter-space saliency profiles for different adversarial attacks on CIFAR-10 and using VGG19 image classifier. Note the magnitude of saliency profiles on y-axis when comparing different attacks.}
    \label{fig:saliency_profiles_cifar10_vgg19}
\end{figure*}
\begin{figure*}[t!]
    \centering
    \subfloat[][Clean]{\includegraphics[width=0.5\textwidth]{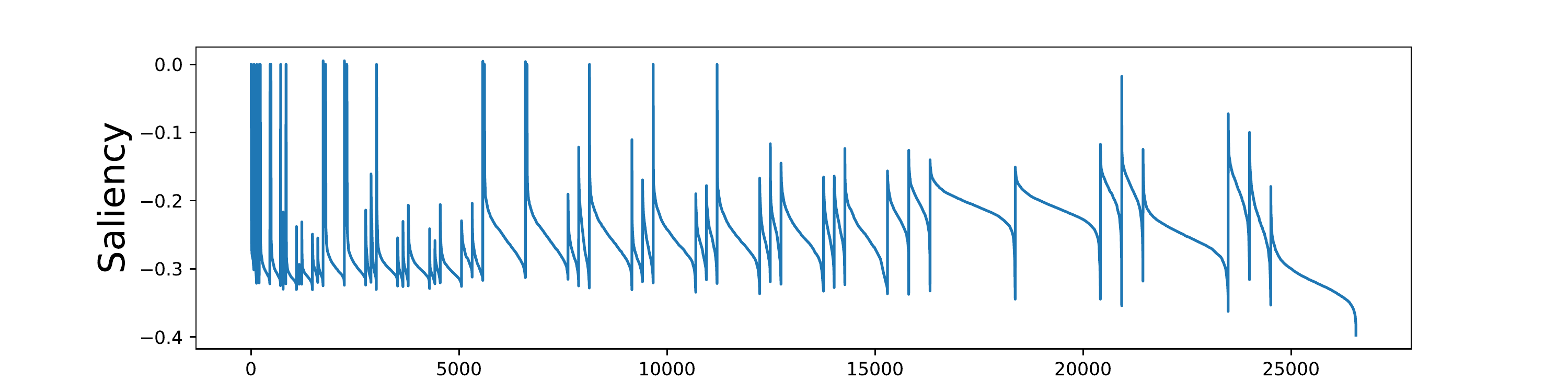}}~
    \subfloat[][CW$L_2$]{\includegraphics[width=0.5\textwidth]{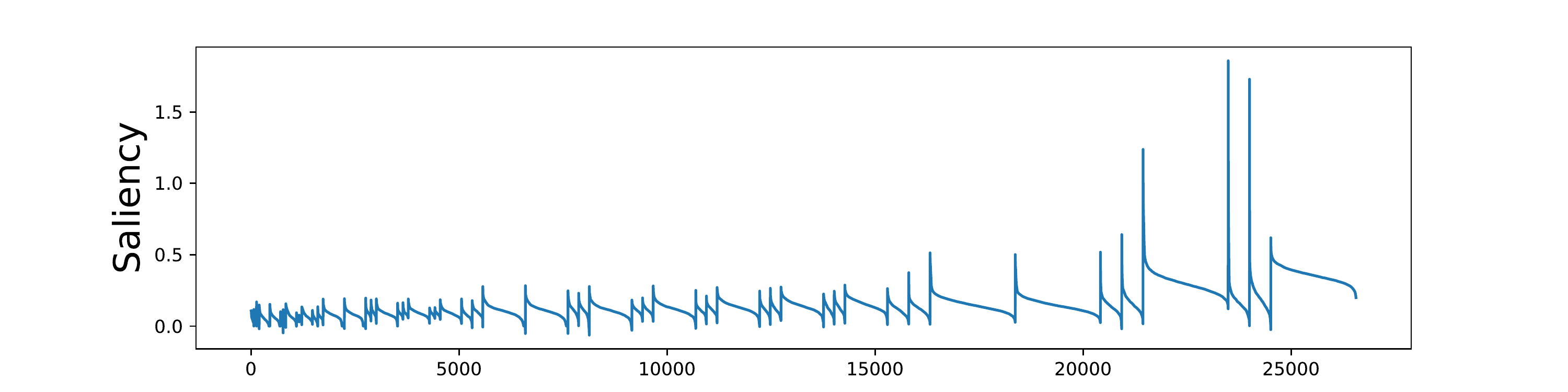}}\\\vspace{-.3cm}
    \subfloat[][CW$L_\infty$]{\includegraphics[width=0.5\textwidth]{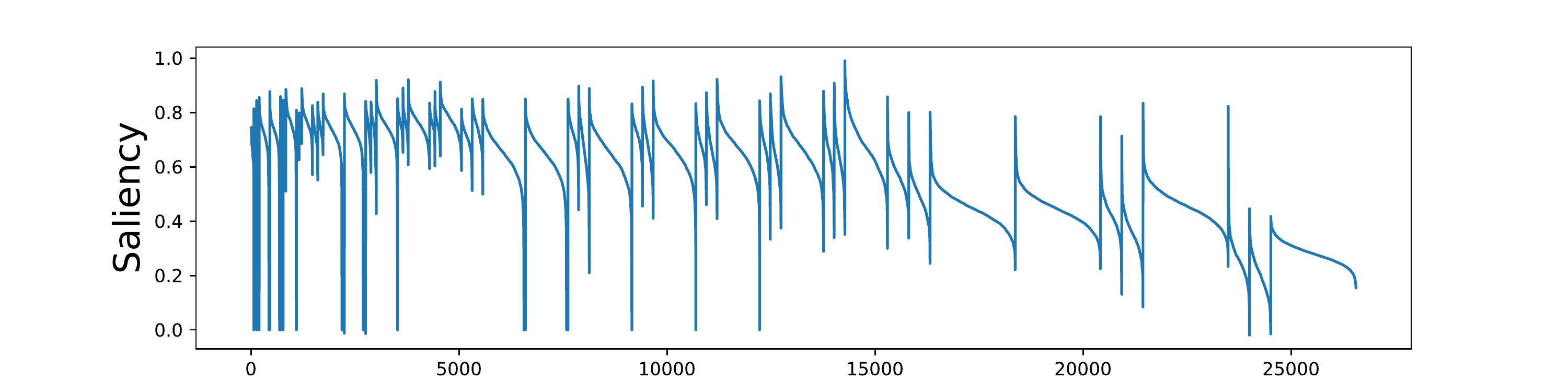}}~
    \subfloat[][PGD]{\includegraphics[width=0.5\textwidth]{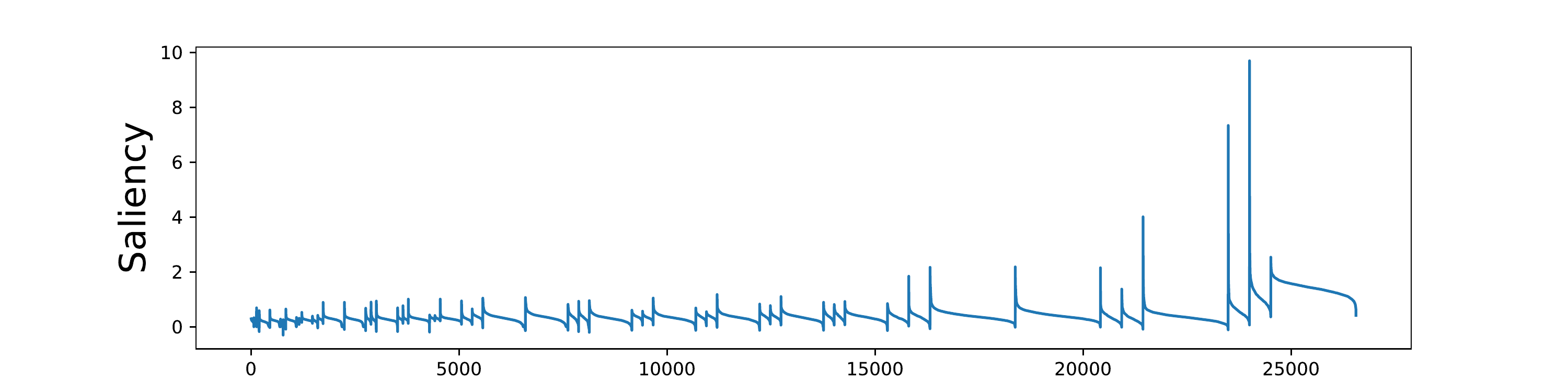}}\\\vspace{-.3cm}
    \subfloat[][Patch]{\includegraphics[width=0.5\textwidth]{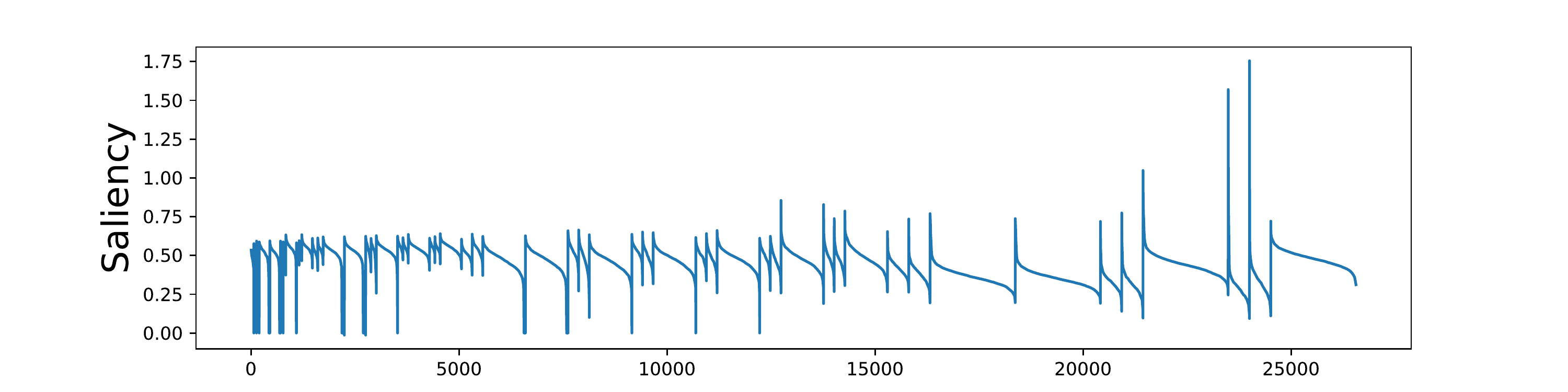}}~
    \subfloat[][DeepFool]{\includegraphics[width=0.5\textwidth]{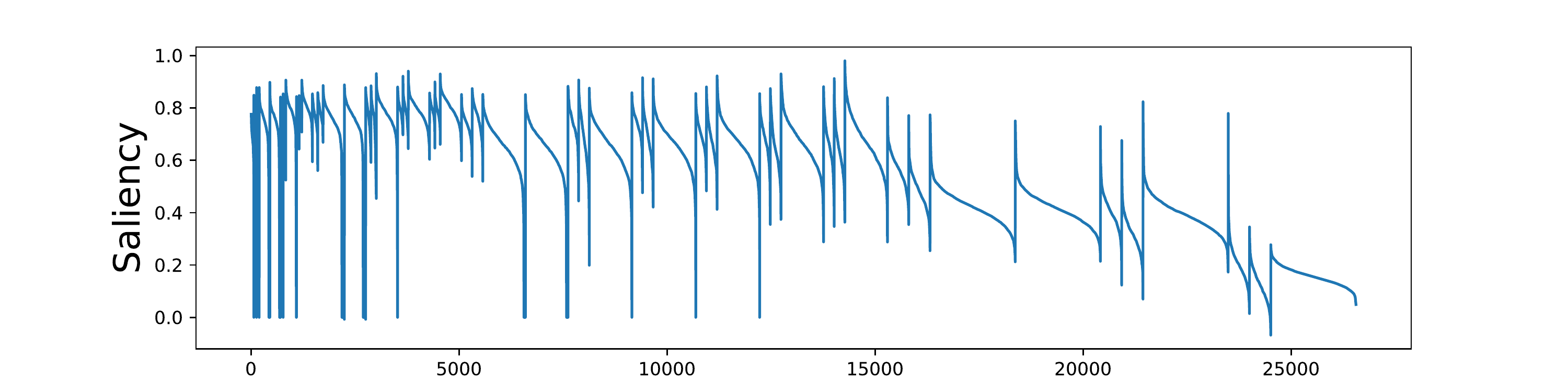}}\\\vspace{0.5cm}
    \caption{Parameter-space saliency profiles for different adversarial attacks on Tiny ImageNet and using ResNet50 image classifier. Note the magnitude of saliency profiles on y-axis when comparing different attacks.}
    \label{fig:saliency_profiles_tiny_resnet50}
\end{figure*}
\begin{figure*}[t!]
    \centering
    \subfloat[][Clean]{\includegraphics[width=0.5\textwidth]{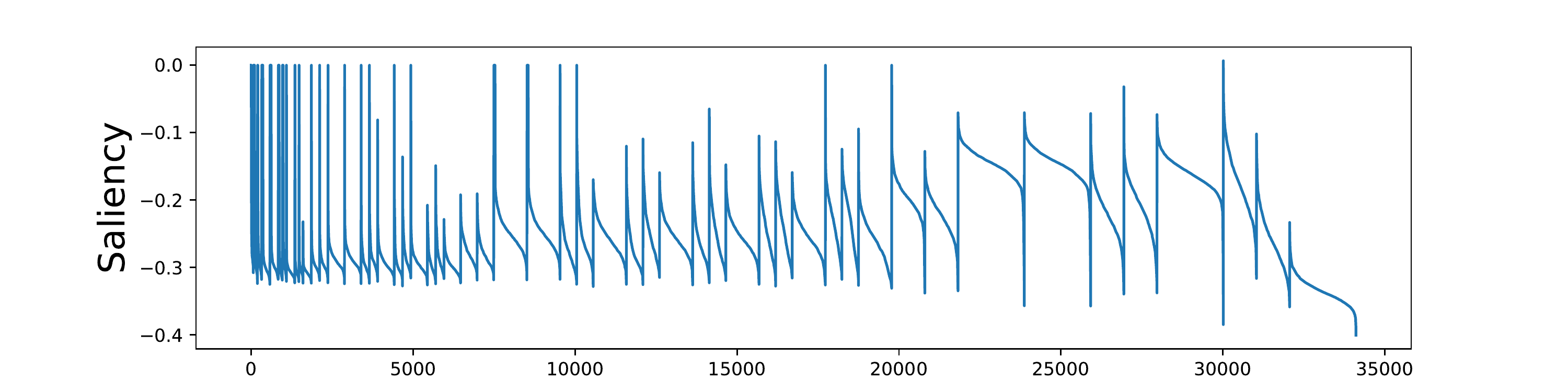}}~
    \subfloat[][CW$L_2$]{\includegraphics[width=0.5\textwidth]{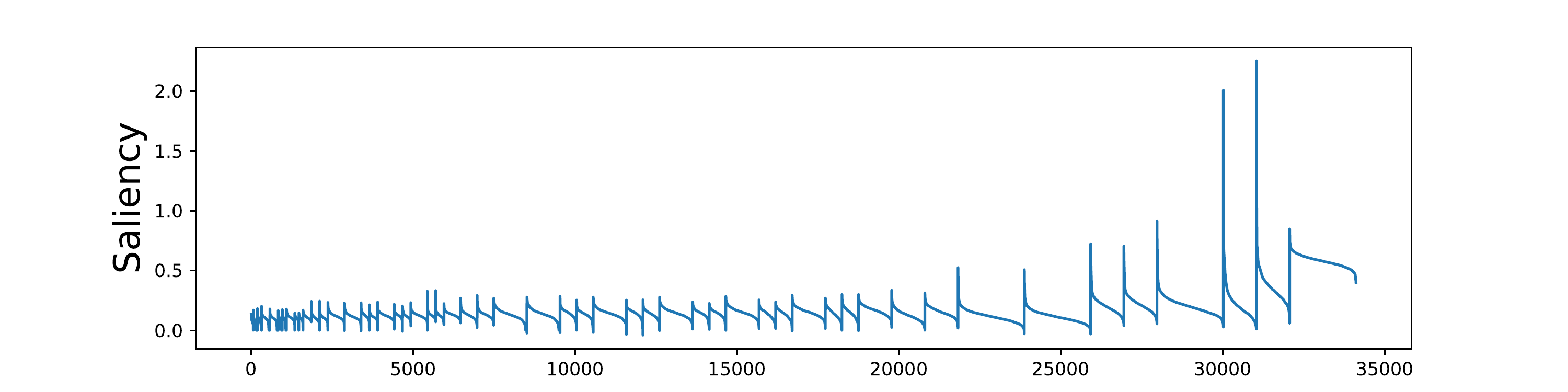}}\\\vspace{-.3cm}
    \subfloat[][CW$L_\infty$]{\includegraphics[width=0.5\textwidth]{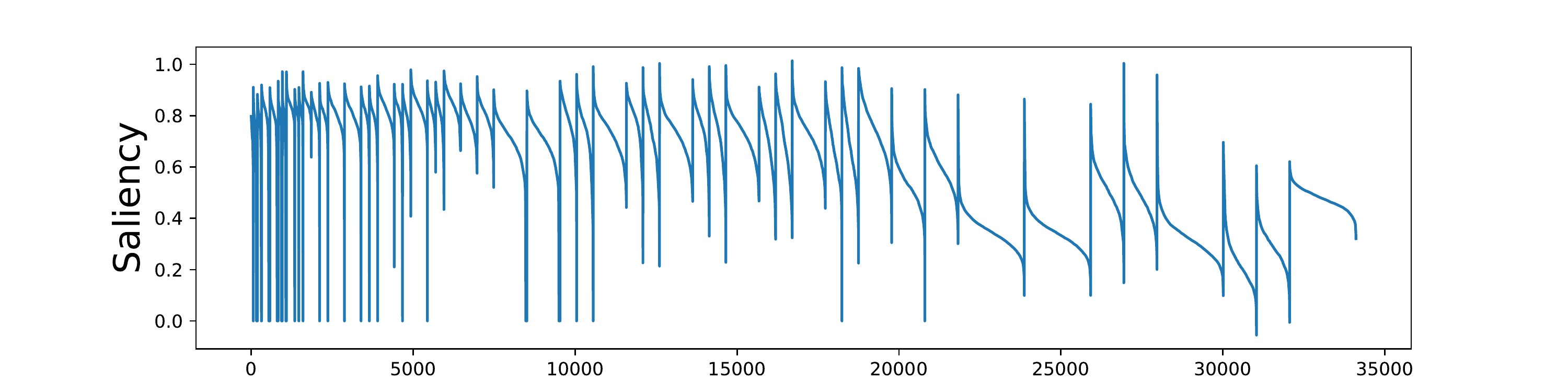}}~
    \subfloat[][PGD]{\includegraphics[width=0.5\textwidth]{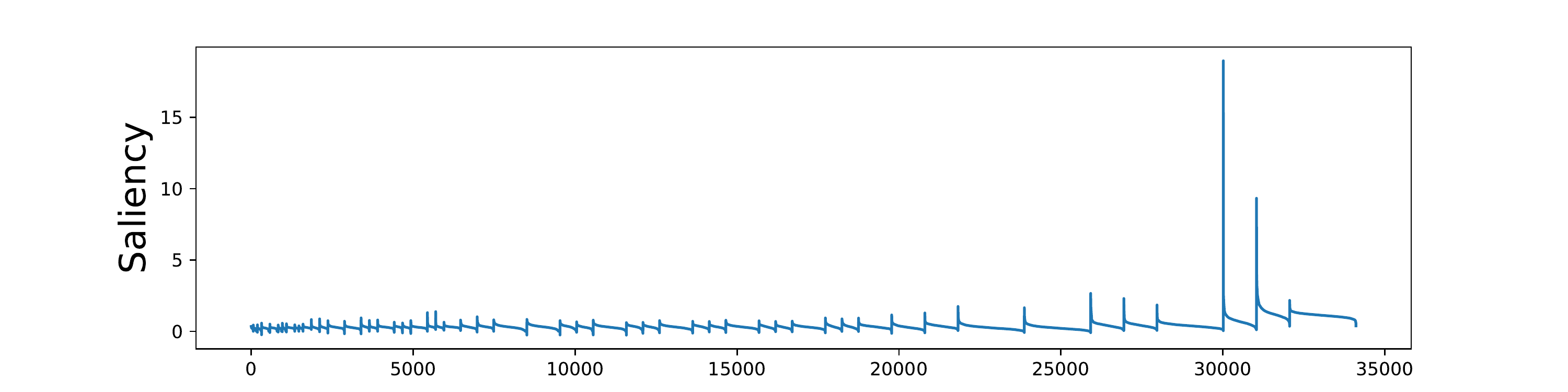}}\\\vspace{-.3cm}
    \subfloat[][Patch]{\includegraphics[width=0.5\textwidth]{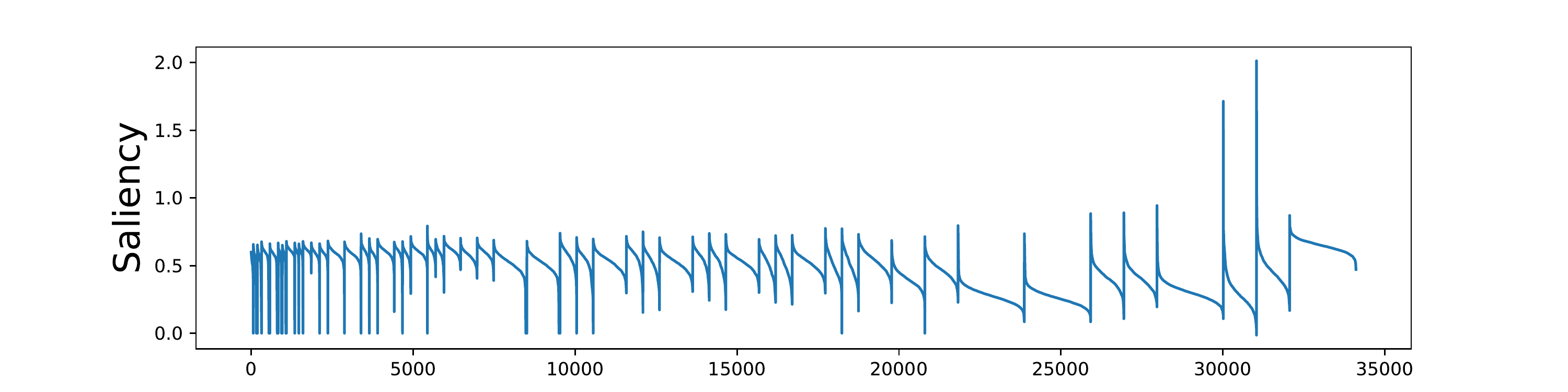}}~
    \subfloat[][DeepFool]{\includegraphics[width=0.5\textwidth]{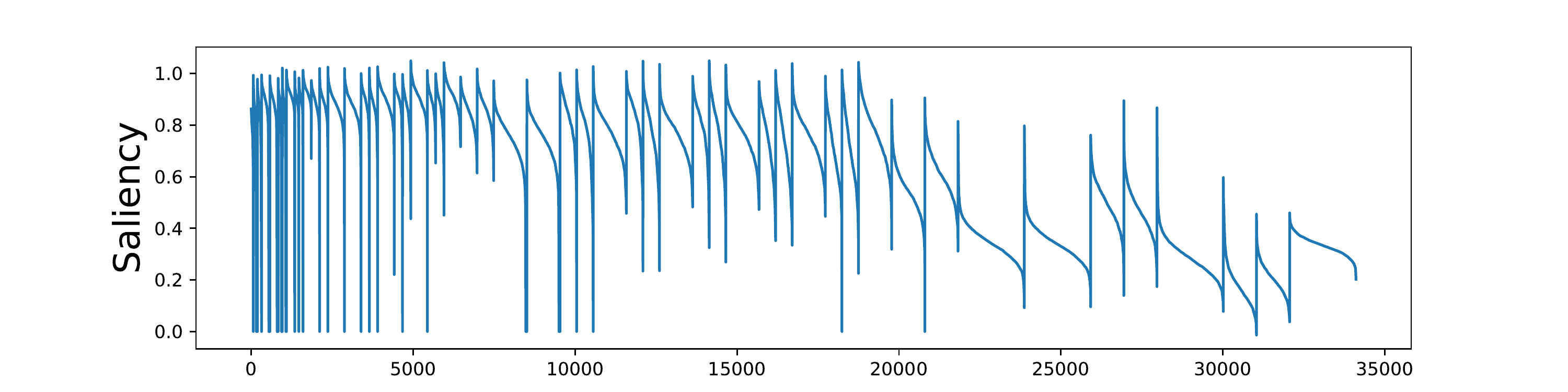}}\\\vspace{0.5cm}
    \caption{Parameter-space saliency profiles for different adversarial attacks on Tiny ImageNet and using ResNext50 image classifier. Note the magnitude of saliency profiles on y-axis when comparing different attacks.}
    \label{fig:saliency_profiles_tiny_resnext50}
\end{figure*}
\begin{figure*}[t!]
    \centering
    \subfloat[][Clean]{\includegraphics[width=0.5\textwidth]{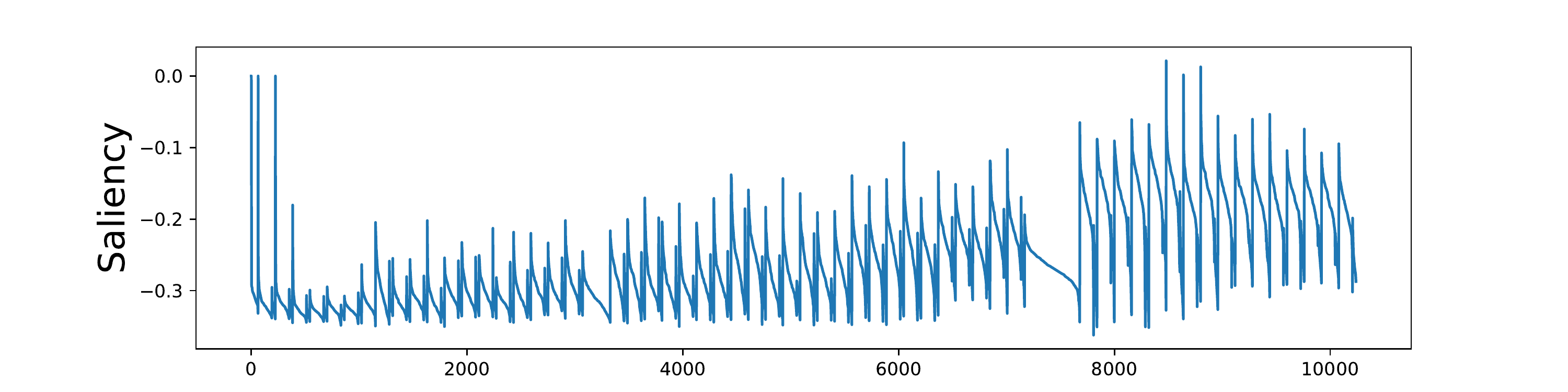}}~
    \subfloat[][CW$L_2$]{\includegraphics[width=0.5\textwidth]{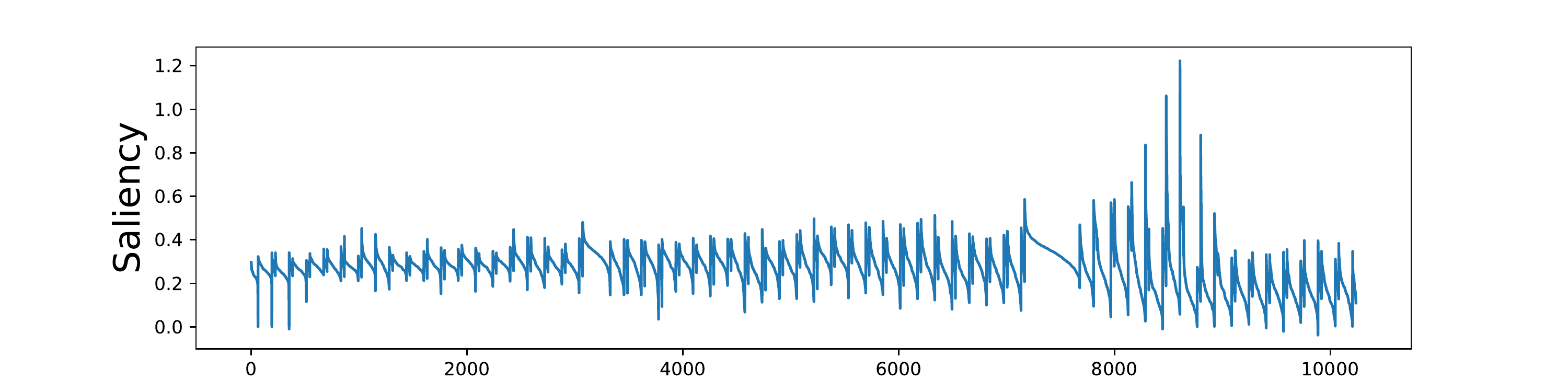}}\\\vspace{-.3cm}
    \subfloat[][CW$L_\infty$]{\includegraphics[width=0.5\textwidth]{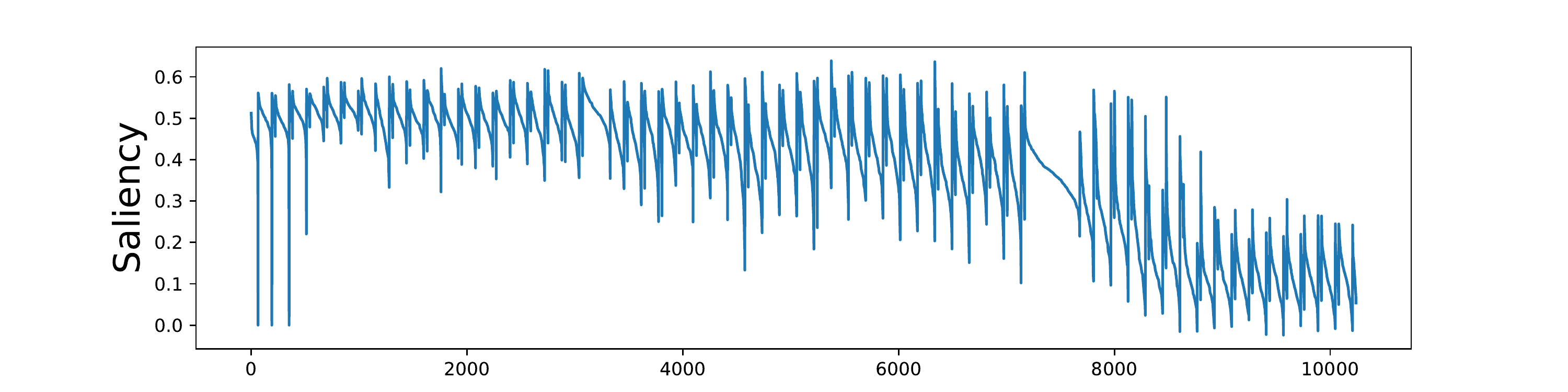}}~
    \subfloat[][PGD]{\includegraphics[width=0.5\textwidth]{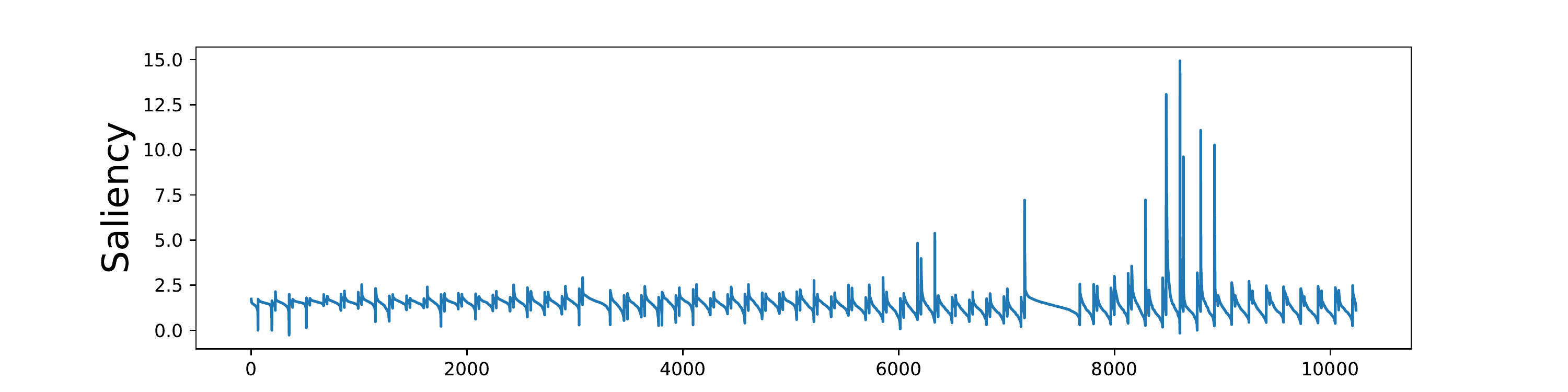}}\\\vspace{-.3cm}
    \subfloat[][Patch]{\includegraphics[width=0.5\textwidth]{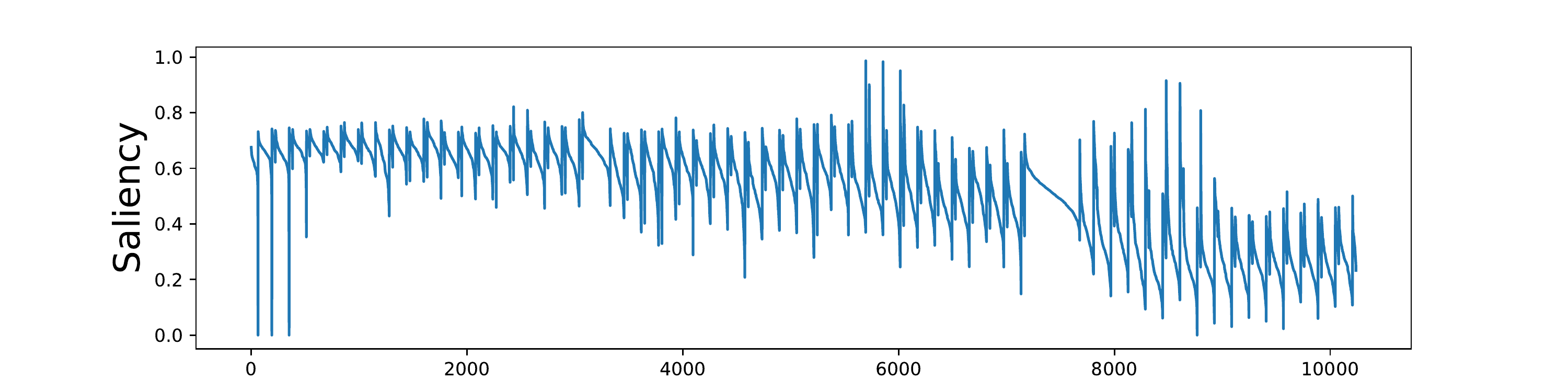}}~
    \subfloat[][DeepFool]{\includegraphics[width=0.5\textwidth]{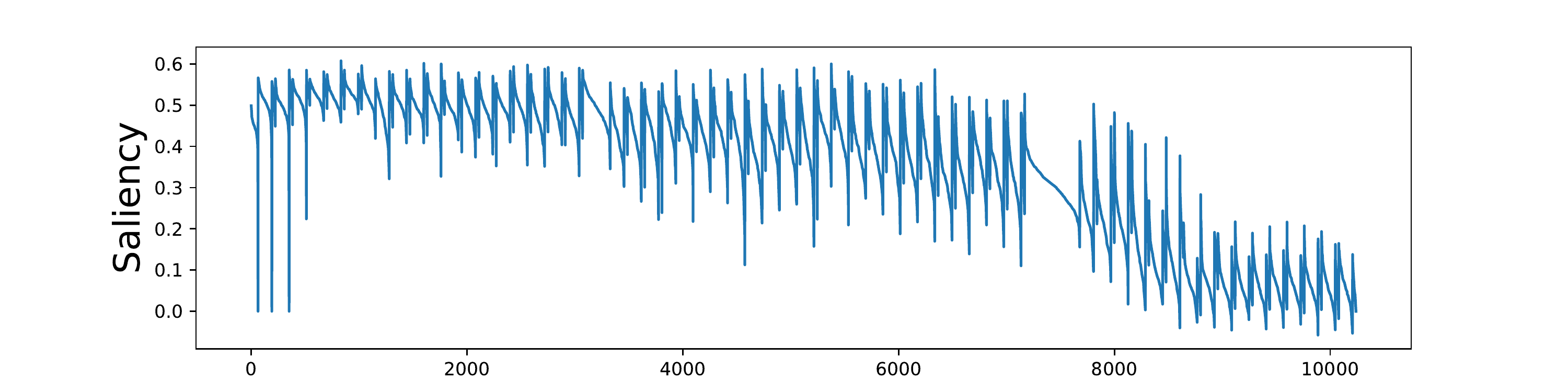}}\\\vspace{0.5cm}
    \caption{Parameter-space saliency profiles for different adversarial attacks on Tiny ImageNet and using DenseNet121 image classifier. Note the magnitude of saliency profiles on y-axis when comparing different attacks.}
    \label{fig:saliency_profiles_tiny_denseset121}
\end{figure*}
\begin{figure*}[t!]
    \centering
    \subfloat[][Clean]{\includegraphics[width=0.5\textwidth]{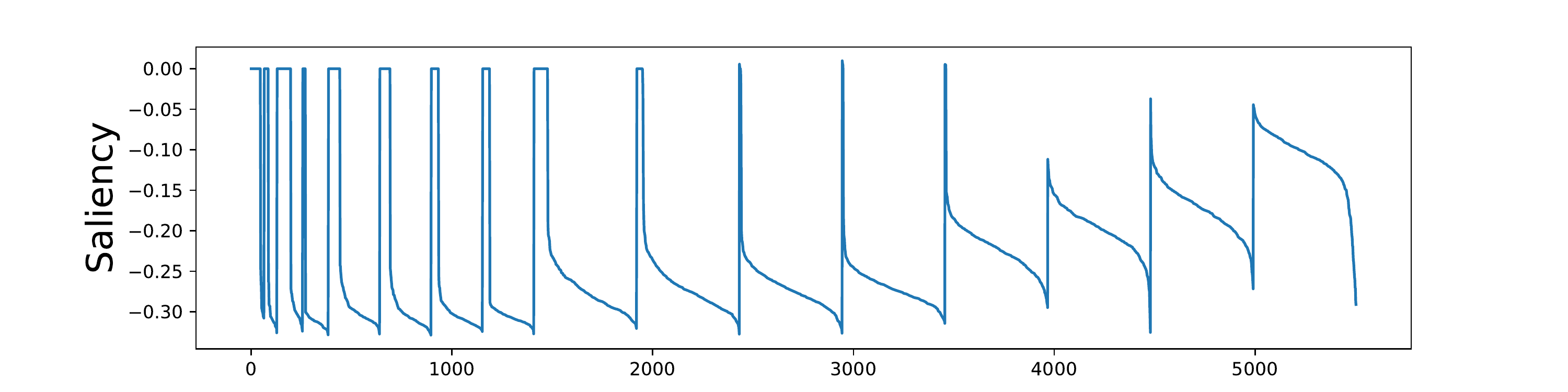}}~
    \subfloat[][CW$L_2$]{\includegraphics[width=0.5\textwidth]{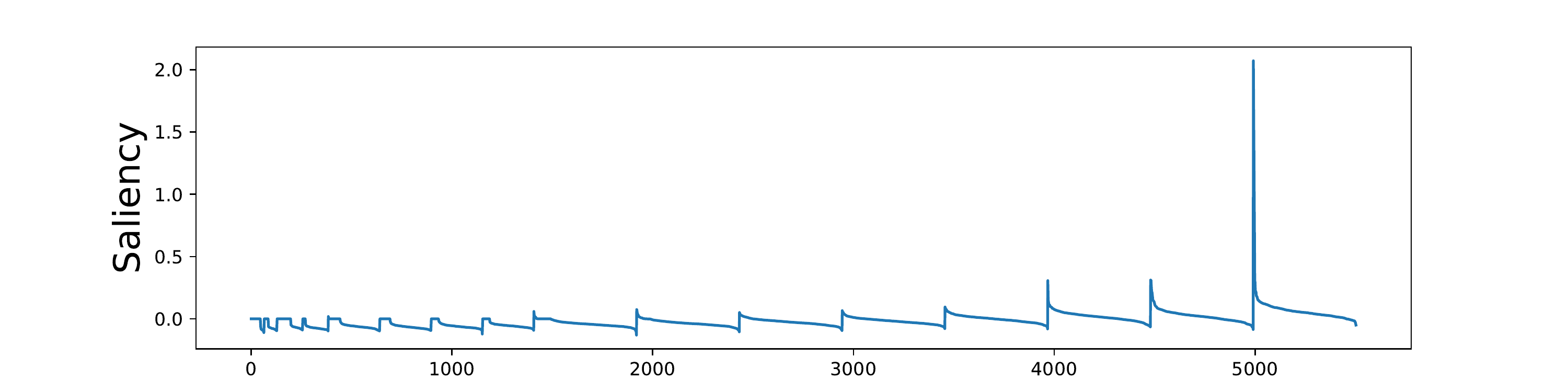}}\\\vspace{-.3cm}
    \subfloat[][CW$L_\infty$]{\includegraphics[width=0.5\textwidth]{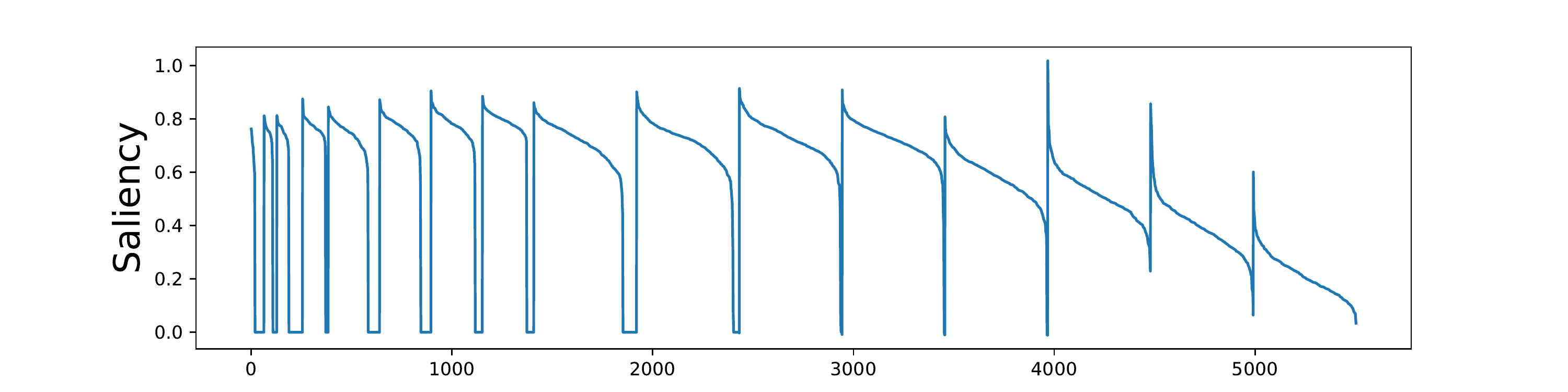}}~
    \subfloat[][PGD]{\includegraphics[width=0.5\textwidth]{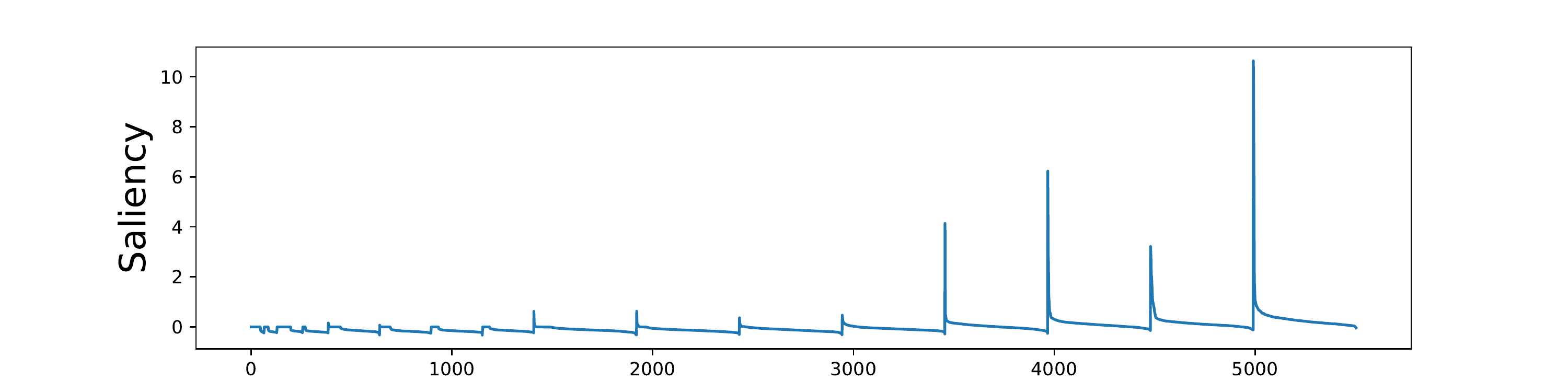}}\\\vspace{-.3cm}
    \subfloat[][Patch]{\includegraphics[width=0.5\textwidth]{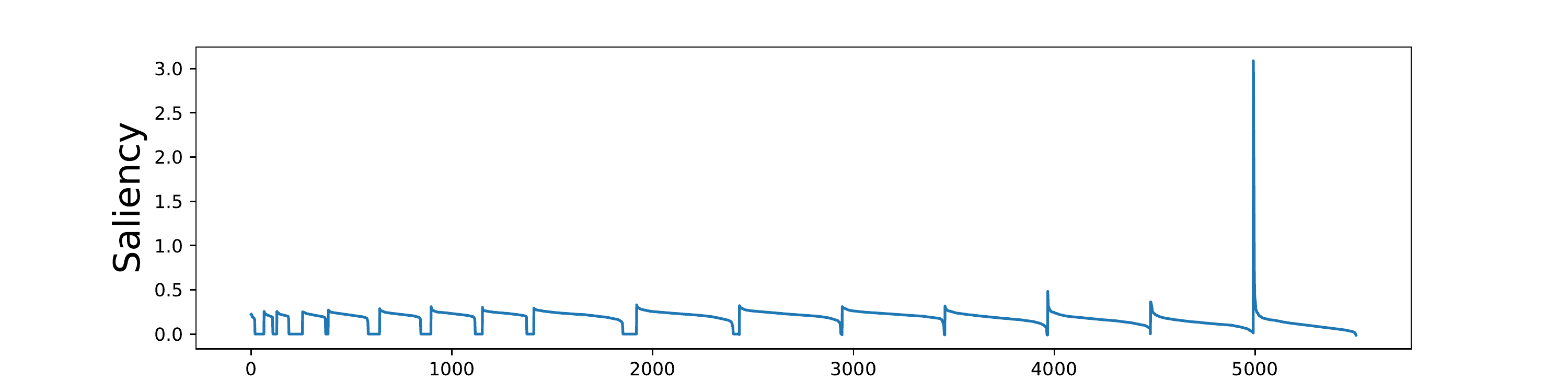}}~
    \subfloat[][DeepFool]{\includegraphics[width=0.5\textwidth]{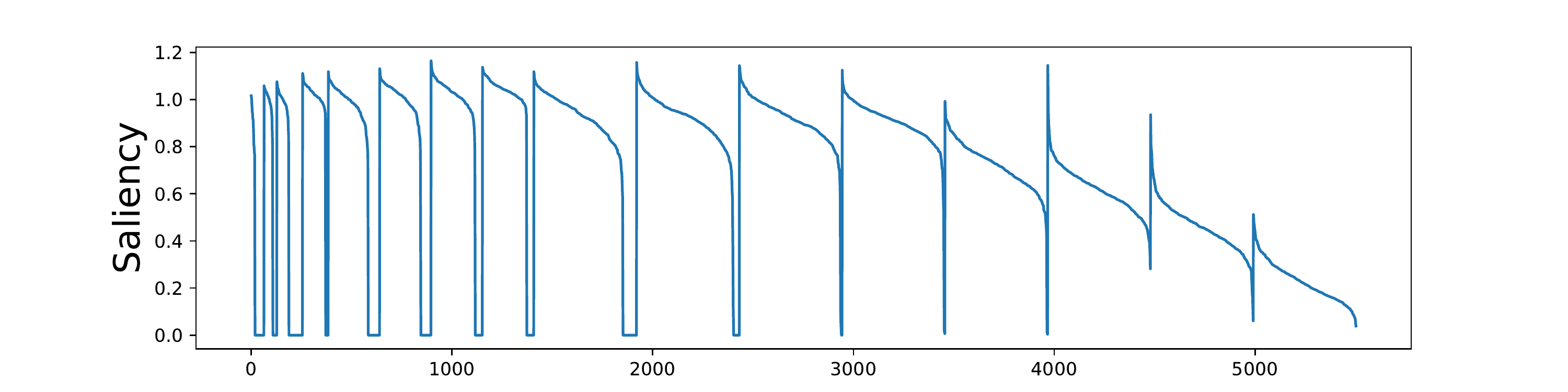}}\\\vspace{0.5cm}
    \caption{Parameter-space saliency profiles for different adversarial attacks on Tiny ImageNet and using VGG19 image classifier. Note the magnitude of saliency profiles on y-axis when comparing different attacks.}
    \label{fig:saliency_profiles_tiny_vgg19}
\end{figure*}

\subsection{Parameter-space Saliency Profile}
In this section we present the averaged parameter-space saliency profile over the test set for CIFAR-10 and Tiny ImageNet dataset and for ResNet50, ResNext50, DenseNet121 and VGG19 image classifiers. Also note that to compute parameter-space saliency profiles, we only use the samples that were originally correctly classified by the image classifier but were miss-classified after being attacked by the attack algorithm. 
\subsubsection{CIFAR-10}
Figures \ref{fig:saliency_profiles_cifar10_resnet50}, \ref{fig:saliency_profiles_cifar10_resnext50}, \ref{fig:saliency_profiles_cifar10_denseset121} and \ref{fig:saliency_profiles_cifar10_vgg19} depict the parameter-saliency profile of different attack algorithms for ResNet50, ResNext50, DenseNet121 and VGG19 image classifiers respectively. Note that the x-axis varies in figures corresponding to different image classification architectures as they have different number of layers and different number of filters for each layer. 

Comparing these figures, clearly shows that the effect of adversarial attacks on victim models is specific to attack algorithms and regardless of the victim model's architecture. In addition, while the saliency profile of CW$L_2$, PGD and Patch attacks seems to be similar, their respective saliency profiles are quite different in their magnitude, \emph{i.e.} the y-axis values.

\subsubsection{Tiny ImageNet}
Figures \ref{fig:saliency_profiles_tiny_resnet50}, \ref{fig:saliency_profiles_tiny_resnext50}, \ref{fig:saliency_profiles_tiny_denseset121} and \ref{fig:saliency_profiles_tiny_vgg19} show the parameter-saliency profile of different attack types for ResNet50, ResNext50, DenseNet121 and VGG19 image classification networks respectively. Similar to CIFAR-10 dataset, the parameter-saliency profile is attack-specific and irrespective of the target victim model. Additionally, while the examination of parameter-saliency profiles for different models may show differences across CIFAR-10 and Tiny ImageNet datasets, it can be observed that there are qualitative similarities which is consistent for each attack algorithm. The cause for cross dataset differences in saliency profiles can be rooted in the fact that Tiny ImageNet dataset is much more diverse and challenging compared to CIFAR-10.

\end{document}